\documentclass{article}

\usepackage[main, final, nonatbib]{neurips_2026}
\usepackage{mathtools}
\usepackage{amssymb}
\usepackage{natbib} 
\usepackage{amsmath}
\usepackage{amsthm}
\usepackage[utf8]{inputenc} 
\usepackage[T1]{fontenc}    
\usepackage{hyperref}       
\usepackage{url}            
\usepackage{booktabs}       
\usepackage{amsfonts}       
\usepackage{nicefrac}       
\usepackage{microtype}      
\usepackage{xcolor}         
\usepackage{algorithm} 
\usepackage{algorithmic}
\usepackage{multirow}
\usepackage{amsthm}
\usepackage{wrapfig}
\usepackage{tcolorbox}
\newcommand{\bm}[1]{\boldsymbol{#1}}
\usepackage{adjustbox}

\newtheorem{theorem}{Theorem}
\newtheorem*{theorem*}{Theorem}
\newtheorem{lemma}[theorem]{Lemma}

\newtheorem{proposition}[theorem]{Proposition}
\title{\textsc{CoupledFlow}: One-Step Neural Operators for Coupled Multi-Physics PDEs  }

\setcitestyle{authoryear,round}

\author{%
   Trong Khiem Tran\footnotemark[1] \\
   Washington State University \\
   \texttt{khiem.tran@wsu.edu} \\
   \And
   Minh Long Bui \\
    Washington State University \\
   \texttt{trongminhlong.bui@wsu.edu} \\
     \And 
     Phi Le Nguyen \\
    Hanoi University of Science and Technology \\
   \texttt{lenp@soict.hust.edu.vn} \\
     \And 
    Mrinal Sen \\
    University of Texas, Austin \\
   \texttt{mrinal@utexas.edu} \\
   \And
   Sanjay Srinivasan \\
   Penn State University \\
   \texttt{szs27@psu.edu} \\
   \And
   Trong Nghia Hoang\footnotemark[1] \\
    Washington State University \\
   \texttt{trongnghia.hoang@wsu.edu} \\
   \And
   Jana Doppa\thanks{Corresponding authors.} \\
    Washington State University \\
   \texttt{jana.doppa@wsu.edu} \\
}

\begin{document}

\maketitle

\begin{abstract}
This paper introduces \textsc{CoupledFlow}, a new neural operator learning framework for multi-physics PDE systems that captures the co-evolution of multiple physical processes by modeling their cross-process interactions.~Unlike existing neural operator methods that primarily target single-physics settings, \textsc{CoupledFlow} employs a novel structure-constrained flow matching algorithm to learn process-specific mean velocity fields conditioned on the full multi-physics state.~This enables fast and accurate one-step prediction of coupled PDE systems that would otherwise require expensive numerical simulations.~A key challenge in training \textsc{CoupledFlow} however lies in sampling structure-constrained conditional flow interpolants that capture cross-process correlations and serve as training examples for flow matching.~To address this challenge, we derive a {\em multi-physics mean-flow consistency identity} that characterizes cross-process interactions and defines the structure-constrained mean-flow matching loss.~To provide informative training samples for flow matching with this loss, we develop a principled method for sampling {\em correlated conditional interpolants based on posterior multi-output Gaussian processes}, which capture richer cross-process correlations than the commonly used endpoint-conditioned linear interpolants.~We also establish a theoretical result showing that the Wasserstein-$2$ distance between the predicted and oracle PDE solution distributions is upper-bounded in terms of the structure-constrained flow matching loss and the smoothness of the learned flow.~Experiments across diverse coupled PDE benchmarks demonstrate that \textsc{CoupledFlow} consistently outperforms existing state-of-the-art neural operators and generative PDE solvers.
\end{abstract}

\section{Introduction}
\label{sec:intro}
Partial differential equations (PDEs) underpin the modeling of physical systems across science and engineering~\citep{turing52the,Gao2020}.~In particular, PDEs describe how physical quantities evolve and interact across space and time through differential equations, which can be solved using numerical simulations.~However, these simulations can be computationally expensive, especially for high-dimensional physical systems.~To mitigate this computational cost, a prominent direction in scientific machine learning is to learn surrogate mappings directly from simulation data that approximate the underlying solution operators. Models that learn such mappings between input and solution function spaces are commonly referred to as neural operators.~Notable state-of-the-art methods include FNO~\citep{li2021fourier}, DeepONet~\citep{lu2021learning} and CViT~\citep{wang2025cvitcontinuousvisiontransformer}. 

{\bf Research Gap.}~Existing neural operators have primarily been developed for single-physics PDEs, whereas many real-world systems involve interacting physical processes that co-evolve and influence one another.~Examples include reaction-diffusion systems in chemistry and biology~\citep{turing52the}, which model non-linear interactions between multiple species, and thermo-hydro-mechanical systems~\citep{Gao2020} where temperature, pressure, and deformation variables are tightly coupled.~In such systems, the future state of each process depends on the current states of others and thus cannot be predicted accurately in isolation. Therefore, developing neural operators that explicitly capture these cross-physics dependencies is important for prediction accuracy.~However, extending single-physics neural operators to coupled systems is challenging.~For example, naively concatenating internal neural representations across different physical processes often leads to poor results~\citep{xiao2023coupled}, as this approach does not explicitly model cross-physics interactions.

More recent work addresses this challenge by introducing explicit cross-physics information exchange into neural operator architectures.~For example, CMWNO~\citep{xiao2023coupled} models coupling through multi-wavelet decomposition, while COMPOL~\citep{sun2025compolunifiedneuraloperator} maintains process-specific FNO streams and exchanges information through recurrent or attention-based aggregation.~Despite their improved performance, these approaches primarily capture cross-physics interactions through architectural design and endpoint regression which largely focuses on mapping initial states to final states without explicitly learning correlations among physical processes along their intermediate transport paths.~This limitation can lead to suboptimal predictive performance (see Section~\ref{sec:experiments}).

{\bf New Perspective.}~To explicitly capture correlations among physical processes along intermediate transport paths, we revisit neural operator learning from the lens of a mass-transport flow.~We view the mapping from initial conditions to PDE solutions as transporting a source distribution over initial states to a target distribution over solutions.~Such transport flow can be learned through conditional flow (FM)~\citep{lipman2022flow,baldan2025flow, utkarsh2025physics} which has recently emerged as a powerful approach for generative modeling.~Importantly, by leveraging mean-flow parameterization~\citep{geng2025mean}, we can learn intermediate transport dynamics while preserving the one-step prediction capability of neural operators.


However, extending FM to coupled multi-physics systems is challenging as it requires accounting for cross-process dependencies in both the parameterized flow and its training examples.~To understand this challenge, we view FM training from a {\em proposer-filter} perspective where (1) conditional interpolants propose transport paths connecting paired initial states and PDE solutions as training examples; and (2) the parameterized flow acts as a filter by imposing consistency constraints on the learned transport.~Since FM amortizes information from these conditional paths into the parameterized flow, constructing structure-constrained interpolants that capture cross-process correlations is critical for effective learning of coupled multi-physics systems.


$\blacktriangleright$ In multi-physics modeling scenarios, two key challenges remain unresolved: (1) how to develop a mean-flow representation that explicitly captures interactions among dependent physical processes; and (2) how to construct correlated intermediate transport paths that provide informative supervision for learning these interactions.~Our experiments show that a vanilla application of FM with independent linear interpolants is often suboptimal in coupled PDE systems (see Section~\ref{sec:experiments}).

{\bf Main Contributions.}~To address these challenges, we make the following technical contributions:

{\bf 1.~Algorithm Design.}~We develop \textsc{CoupledFlow}, a one-step multi-physics neural operator built on structure-constrained mean-flow matching to explicitly learn cross-process interactions.~In particular, we derive a {\em multi-physics mean-flow consistency identity} that characterizes the coupling among process-specific transport flows.~This identity enables us to impose structural constraints on the parameterized multi-physics mean-flow representation by penalizing constraint violations during flow matching (Section~\ref{subsec:identity}).~To provide informative training samples for the structure-constrained flow matching loss, we develop a principled method for sampling {\em correlated conditional interpolants} from posterior multi-output Gaussian processes, which capture richer cross-process correlations than conventional endpoint-conditioned linear interpolants(Section~\ref{subsec:gp}).

{\bf 2.~Theoretical Analysis.}~We characterize test-time prediction performance in terms of the Wasserstein-2 distance between predicted and oracle PDE solution distributions and show that it is upper-bounded in terms of the coupled mean-flow matching loss (see Theorem~\ref{thm:2}, Section~\ref{subsec:algo}).~This analysis also clarifies how the smoothness of the learned transport flow influences prediction error.~We further empirically evaluate the resulting error bounds for correlated GP-based interpolants and standard linear interpolants (see Section~\ref{sec:experiments}).

{\bf 3.~Empirical Evaluation.}~We evaluate \textsc{CoupledFlow} on diverse coupled PDE benchmarks spanning different physical systems.~The results demonstrate significant improvements in prediction accuracy over existing (coupled) neural operators and generative PDE solvers.~We also conduct ablation studies to isolate the contributions of the coupled mean-flow representation and GP-based correlated interpolants, demonstrating the importance of both components for accurate multi-physics prediction (see Section~\ref{sec:experiments}).

\section{Background and Problem Setup}\label{sec:background}
{\bf A.~Coupled Physics Systems.}~We consider a multi-physics system consisting of $m$ interacting physical processes $\{\bm{u}_i(\bm{x}, t)\}_{i=1}^m$ modeled by coupled PDEs.~Each PDE can be characterized as:
\begin{eqnarray}
\hspace{-4mm}\bm{u}_r\big(\bm{x},\tau\big) \hspace{-2mm}&\triangleq&\hspace{-2mm} T_r\Big[\bm{x} \mid \mathcal{J}^{\kappa}\bm{u}_1(\bm{x}), \ldots, \mathcal{J}^{\kappa}\bm{u}_m(\bm{x})\Big] \ \text{where}\ \mathcal{J}^{\kappa}\bm{u}_i(\bm{x}) \triangleq \Big\{\partial_{\bm{x}}^p\partial_t^q\bm{u}_i(\bm{x}) \mid p + q \leq \kappa\Big\} , \label{eq:1}
\end{eqnarray}
Here, $T_r$ maps from the initial spatiotemporal condition $\bm{x}$ at time $t = 0$ to the corresponding solution $\bm{u}_r(\bm{x}, \tau)$ at the target time $t = \tau$.~As each operator $T_r$ depends on differential information from other processes, each individual process $\bm{u}_r(\bm{x}, t)$ cannot be solved in isolation.~Furthermore, forward simulation of the coupled system can be computationally expensive.~This motivates learning a data-driven approximation of the operator $\bm{u}_r(\bm{x}, \tau) = T_r[\bm{x}\mid \ldots]$ using an efficient one-step predictor $\bm{u}_r(\bm{x}, \tau) \simeq \bm{u}_r(\bm{x}, \tau; \theta)$, thus avoiding repeated numerical simulation of $T_r$.

Existing neural operators differ in how they parameterize the surrogate $\bm{u}_r(\bm{x}, \tau; \theta)$ and learn its parameters $\theta$.~However, most approaches learn to map between initial conditions and corresponding simulation outputs during training~\citep{li2021fourier, lu2021learning, huang2024diffusionpde} without explicitly modeling how interactions among physical processes drive the co-evolution of their intermediate states.~For multi-physics systems, such interactions play a central role in determining the solution, motivating a learning framework that explicitly captures cross-process dependencies.

{\bf B.~Flow-based Neural Operator.}~To explicitly model intermediate transport dynamics,, we can view the solution operator $T_r$ as transporting a source distribution over initial conditions $\bm{x}$. The target distribution is implicitly defined as the push-forward of the source distribution $\pi_s$ through $T_r$.
~We can then parameterize this transport using an ODE flow $\mathrm{d}\bm{x}(t)/\mathrm{d}t = \bm{u}(\bm{x}(t), t) \  \simeq\ \bm{u}_\theta(\bm{x}(t), t)$.

The neural velocity field $\bm{u}_\theta(\bm{x}(t), t)$ can be learned using conditional linear interpolants $\bm{x}(t \mid \bm{x}, \bm{x}') = (1 - t/\tau)\bm{x} + (t/\tau)\bm{x}'$\footnote{As we consider $t \in [0,\tau]$, this construction guarantees that the interpolation reaches $\bm{x}'$ when $t = \tau$.} and amortizing them into $\bm{u}_\theta(\bm{x}(t), t)$~\citep{lipman2022flow}:
\begin{eqnarray}
\hspace{-30mm}\theta &\triangleq& \arg\min_\theta\ \mathbb{E}_{t\in \mathcal{U}(0,\tau)}\mathbb{E}_{\bm{x}\sim \pi_s}\mathbb{E}_{\bm{x}' \sim \pi_\tau}\big\|\bm{u}_\theta(\bm{x}(t), t) - \mathrm{d}\bm{x}(t \mid \bm{x}, \bm{x}')/\mathrm{d}t\big\|^2 \ ,\label{eq:3}
\end{eqnarray}
where $\mathrm{d}\bm{x}(t \mid \bm{x}, \bm{x}')/\mathrm{d}t = (1/\tau)(\bm{x}' - \bm{x})$.~To enable one-step prediction during test time, the above can be further empowered with a mean flow parameterization~\citep{geng2025mean} as detailed next.

{\bf C.~Mean-Flow Neural Operator.}~We will use a direct parameterization for the mean or average velocities $\bar{\bm{u}}_\theta(\bm{x}(t), s, t) \simeq \bar{\bm{u}}(\bm{x}(t), s, t)$ over interval $[s, t]$ where
\begin{eqnarray}
\hspace{-26.5mm}\bar{\bm{u}}_\theta(\bm{x}(t), s, t) &=& (t - s)^{-1}\int_s^t \bm{u}_\theta(\bm{x}(\varepsilon), \varepsilon)\mathrm{d}\varepsilon \ \ = \ \ (t - s)^{-1}(\bm{x}(t) - \bm{x}(s)) \ , \label{eq:4}      
\end{eqnarray}
via the fundamental theorem of calculus.~By slight abuse of notation, the mean flow is parameterized as a function of the current state and time interval.~Differentiating both sides w.r.t $t$, we obtain:
\begin{eqnarray}
\hspace{-46.5mm}\bm{u}_\theta(\bm{x}(t), t) &=& \bar{\bm{u}}_\theta(\bm{x}(t), s, t) \ \ +\ \  (t - s) \cdot \mathrm{d}\bar{\bm{u}}_\theta(\bm{x}(t), s, t)/\mathrm{d}t \ ,\label{eq:5}
\end{eqnarray}
which means we can either parameterize the instantaneous velocity $\bm{u}_\theta(\bm{x}(t), t)$ directly or compute it via the parameterization of the mean velocities $\bar{\bm{u}}_\theta(\bm{x}(t), s, t)$.~The latter preserves the flow structure while enabling fast one-step inference during test time.~It can be learned via plugging Eq.~\eqref{eq:5} into Eq.~\eqref{eq:3} and taking expectation over $s \sim \mathcal{U}(0, \tau)$ and $t\sim \mathcal{U}(s,\tau)$~\citep{geng2025mean}. 

{\bf D.~Multi-Physics Challenge.}~At a conceptual level, the mean-flow representation admits a rich path-level structure that extends beyond direct matching between source and target states.~In particular, the mean-flow relation in Eq.~\eqref{eq:4} constrains how the average velocity evolves along admissible transport paths, suggesting that the representation has the capacity to capture how multiple physical processes co-evolve in coupled systems.~However, existing mean flow formulations~\citep{geng2025mean} are developed almost exclusively for single-physics settings where the main goal is to enable one-step inference.~Consequently, despite its structural capacity, the representation is ultimately learned via amortizing over conditional linear interpolations between source and target samples as in Eq.~\eqref{eq:3}, leaving its path-level learning potential largely underutilized.

This is sufficient in single-physics settings where the only requirement on the interpolation is that it delivers a valid transport from source to target.~In multi-physics settings, however, we need the transport to reflect how different components of the state are correlated along the path.~The linear interpolation $\bm{x}(t \mid \bm{x}, \bm{x}') = (1-t/\tau)\bm{x} + (t/\tau)\bm{x}'$ unfortunately offers no mechanism for this.~It produces intermediate states as a coordinate-wise function of endpoints alone, leaving no room to capture any structural relationship among components of $\bm{x}(t)$.~To address this limitation, we develop a coupled mean flow construction in which the linear interpolation is replaced by a correlated interpolation that acts as a learning device for proposing viable cross-physics interaction structures, which are subsequently amortized into the neuralized mean flow field (see Section~\ref{sec:method}).

\section{\textsc{CoupledFlow}: Learning One-Step Multi-Physics Neural Operators}
\label{sec:method}

\begin{figure}
    \centering
    \includegraphics[width=\linewidth]{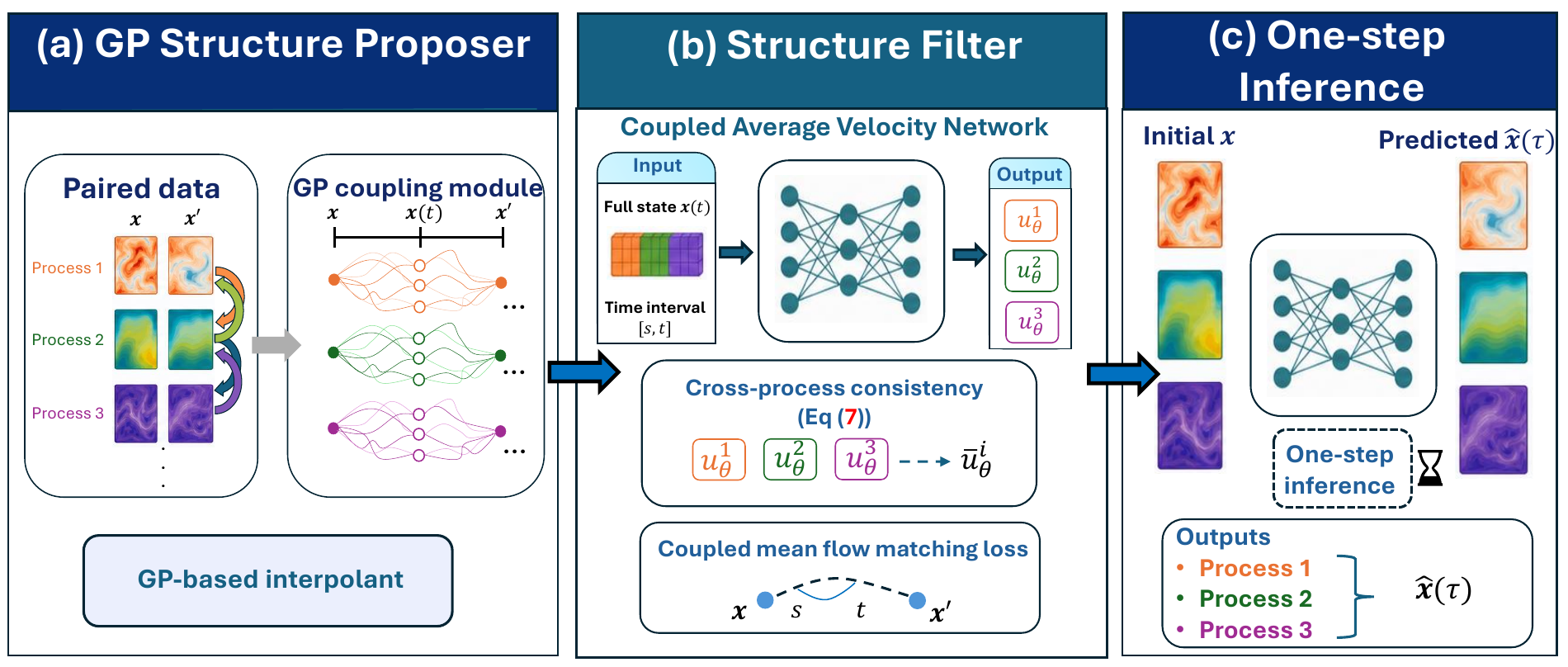}\vspace{-3mm}
    \caption{Workflow diagram of the proposed \textsc{coupled-flow} framework which consists of three main components: (a) a structure proposer that constructs Gaussian process-based conditional flow interpolants to capture cross-physics dependencies (Section \ref{subsec:gp}); (b) a structure filter that constrains coupled transport dynamics through the mean-flow consistency identity (Section \ref{subsec:identity}); and (c) a one-step inference procedure for efficient multi-physics prediction at testing time (Section \ref{subsec:algo}).}\vspace{-6mm}
    \label{fig:CoupledFlow}
\end{figure}


{\bf High-Level Overview of \textsc{CoupledFlow}.}~This section develops a coupled mean-flow framework for one-step multi-physics neural operators.~A key starting point is the mean-flow identity in Eq.~\eqref{eq:4}, which can be expanded to derive a multi-physics mean-flow consistency relation that constrains interactions among different physical processes along transport paths.~The overall framework consists of two complementary components:~(1) a structure proposer that generates candidate cross-physics interaction patterns through structure-constrained flow interpolations, and (2) a structure filter that uses the multi-physics consistency relation to constrain these patterns during learning (see Fig.~\ref{fig:CoupledFlow}).

That is, the correlated interpolations provide learning signals for the coupled mean-flow matching loss during training.~The consistency identity on the other hand constrains the learned mean-flow representation.~At inference time, the learned mean-flow operator can be used to predict the coupled multi-physics state in a single step without explicitly simulating the entire transport flow.

The remainder of this section makes each of these steps concrete.~Section~\ref{subsec:identity} derives the multi-physics consistency identity.~Section~\ref{subsec:gp} constructs the multi-output GP-based interpolation.~Finally, Section~\ref{subsec:algo} describes the high-level algorithm with practical training methods.

\subsection{Structure Filter: Multi-Physics Mean Flow Consistency Identity}
\label{subsec:identity}
Let $\bm{x}(t) = (\bm{x}_1(t), \bm{x}_2(t),\ldots, \bm{x}_m(t))$ denote a multi-physics state comprising the  states of $m$ coupled physical processes at time $t$.~Let $\bar{\bm{u}}_\theta^i(\bm{x}(t), s, t)$ denote the mean flow representation for the $i$-$\text{th}$ physical process over the time interval $[s, t]$.~The dependence among these physical processes is captured by conditioning each individual mean flow representation on the entire multi-physics state.~The corresponding instantaneous velocity at time $t$ can then be obtained from Eq.~\eqref{eq:5} as follows:
\begin{eqnarray}
\hspace{-46mm}\bm{u}_\theta^i\big(\bm{x}(t), t\big) &=& \bar{\bm{u}}_\theta^i\big(\bm{x}(t), s, t\big) \ \ +\ \ (t - s) \cdot \mathrm{d}\bar{\bm{u}}_\theta^i\big(\bm{x}(t), s, t\big)/\mathrm{d}t \ .\label{eq:6}
\end{eqnarray}
Applying the chain rule to the derivative term on the RHS of Eq.~\eqref{eq:6} gives:
\begin{eqnarray}
\hspace{-24.5mm}\mathrm{d}\bar{\bm{u}}_\theta^i\big(\bm{x}(t), s, t\big)/\mathrm{d}t &=& \partial_t\bar{\bm{u}}_\theta^i\big(\bm{x}(t), s, t\big) + \sum_{c=1}^m\bm{u}_\theta^c\big(\bm{x}(t), t\big)\partial_{\bm{x}_c(t)}\bar{\bm{u}}_\theta^i\big(\bm{x}(t), s, t\big) \ . \label{eq:7}
\end{eqnarray}
Substituting Eq.~\eqref{eq:7} into Eq.~\eqref{eq:6} yields the \colorbox{green!20}{{\bf multi-physics consistency identity}},
\begin{eqnarray}
\hspace{-4mm}\bm{u}_\theta^i\big(\bm{x}(t), t\big) \hspace{-2.5mm}&=&\hspace{-2.5mm} \bar{\bm{u}}_\theta^i\big(\bm{x}(t), s, t\big) \hspace{-0.5mm}+\hspace{-0.5mm} (t - s) \Bigg[\partial_t\bar{\bm{u}}_\theta^i\big(\bm{x}(t), s, t\big) + \sum_{c=1}^m\bm{u}_\theta^c\big(\bm{x}(t), t\big)\partial_{\bm{x}_c(t)}\bar{\bm{u}}_\theta^i\big(\bm{x}(t), s, t\big)\Bigg] \label{eq:8}    
\end{eqnarray}

Intuitively, this means that a learned mean-flow representation that accurately models the coupled PDE dynamics must satisfy this constraint.~Therefore, the consistency identity can be used as a structure filter in conditional flow matching.

More specifically, this filter constrains how the cross-physics interaction patterns proposed by endpoint-conditioned interpolations are incorporated into the learned mean-flow representation.
~Interaction patterns consistent with Eq.~\eqref{eq:8} can be captured by the learned mean-flow representation, while inconsistent patterns are constrained by the coupled mean-flow matching objective.
As a consequence, the learned representation captures cross-physics dependencies while satisfying the mean-flow consistency identity.~Mathematically, the classical conditional flow matching approach~\cite{lipman2022flow} can be generalized as follows:
\begin{eqnarray}
\hspace{-21mm}\theta &=& \arg\min_\theta \frac{1}{m}\sum_{i=1}^m\mathbb{E}_{s,t} \mathbb{E}_{(\bm{x}, \bm{x}')} \mathbb{E}_{\bm{x}_i(t)}\Big\|\bm{u}_\theta^i\big(\mathbb{E}[\bm{x}(t)],t\big) \ \ -\ \  \mathrm{d}\bm{x}_i(t\mid \bm{x},\bm{x}')/\mathrm{d}t\Big\|^2 \ ,\label{eq:9}
\end{eqnarray}
Unlike the original loss in Eq.~\eqref{eq:3}, we do not parameterize $\bm{u}_\theta^i(\bm{x}(t), t)$ directly but instead use the form induced by the mean-flow representation in Eq.~\eqref{eq:8}.~This introduces an additional time variable $s \in [0,\tau]$ over which we average the loss in Eq.~\eqref{eq:9}.~The next key component of our framework is the construction of multi-physics interpolations that capture cross-physics dependencies and provide learning signals for the coupled mean-flow matching loss, as detailed in Section~\ref{subsec:gp} below.


\subsection{Structure Proposer: Conditional Flow Interpolation Construction}
\label{subsec:gp}
As discussed in Section~\ref{subsec:identity}, the mean-flow consistency identity in Eq.~\eqref{eq:8} acts as a structural filter that constrains the cross-physics interaction patterns learned from candidate flow interpolants.~The effectiveness of the framework therefore depends on the quality of the endpoint-conditioned flow interpolations $\bm{x}_i(t \mid \bm{x}, \bm{x}')$ for each process $i$, which propose how the intermediate states of different physical processes co-evolve along the transport path.~As such, the interpolations must be capable of generating intermediate coupled states that capture dependencies among multiple processes.

A common choice in conditional flow matching~\citep{lipman2022flow} is the deterministic linear interpolation $\bm{x}_i(t \mid \bm{x}, \bm{x}') = (1 - t/\tau)\bm{x}_i + (t/\tau)\bm{x}'_i $ which reaches $\bm{x}'_i$ when $t=\tau$\footnote{Here, $\bm{x}_i$ and $\bm{x}_i'$ denote the states of process $i$ corresponding to the multi-physics states $\bm{x}$ and $\bm{x}'$, respectively.}.~Although commonly used in single-physics settings, this construction can be restrictive in coupled systems because the intermediate states of different processes are determined by their endpoints without explicitly modeling cross-physics dependencies along the transport path.~As a result, it may fail to capture important cross-physics correlations, limiting the effectiveness of the structural filtering mechanism in Eq.~\eqref{eq:8}. To address this limitation, we replace the independent linear interpolation with a {\em correlated interpolation induced from a multi-output Gaussian process (GP) posterior} conditioned on the endpoint pair $(\bm{x}, \bm{x}')$.~In particular, we assume that the multi-physics path: 
\begin{eqnarray}
\hspace{-44.5mm}\bm{x}(t\mid\bm{x},\bm{x}') &\triangleq& (\bm{x}_1(t\mid\bm{x},\bm{x}'), \bm{x}_2(t\mid\bm{x},\bm{x}'), \ldots, \bm{x}_m(t\mid\bm{x},\bm{x}')) \ , \label{eq:11}
\end{eqnarray}
is stochastic and modeled using a multi-output GP~\citep{alvarez2012kernelsvectorvaluedfunctionsreview} which provides a principled framework for capturing path-level correlations among dependent spatiotemporal processes.~For two dependent processes $(\bm{x}_i(t\mid\bm{x},\bm{x}'),\bm{x}_c(t'\mid\bm{x},\bm{x}'))$, the joint Gaussian distribution is given by:
\begin{eqnarray}
\hspace{-16.5mm}\begin{bmatrix}
\bm{x}_i(t\mid\bm{x},\bm{x}') \\
\bm{x}_c(t'\mid\bm{x},\bm{x}')
\end{bmatrix}
&\sim&
\mathbb{N}
\left(
\begin{bmatrix}
\bm{m}_i(t\mid\bm{x},\bm{x}')\\
\bm{m}_c(t'\mid \bm{x},\bm{x}')
\end{bmatrix},
\begin{bmatrix}
\bm{K}_{ii}(t,t'\mid \bm{x},\bm{x}') & \bm{K}_{ic}(t,t'\mid \bm{x},\bm{x}')\\
\bm{K}_{ci}(t,t'\mid \bm{x},\bm{x}') & \bm{K}_{cc}(t,t'\mid \bm{x},\bm{x}')
\end{bmatrix}
\right)\ , \label{eq:12}
\end{eqnarray}
whose mean and kernel functions determine the marginal correlation structure between these processes.~At each time step $t$, we can thus generate the correlated state $\bm{x}_i(t\mid \bm{x}, \bm{x}')$ given $\bm{x}_c(t \mid \bm{x}, \bm{x}')$ via constructing the posterior GP marginal:
\begin{eqnarray}
\hspace{-59mm}\bm{x}_i(t\mid \bm{x}, \bm{x}')
\mid
\bm{x}_c(t\mid \bm{x}, \bm{x}')
&\sim&
\mathbb{N}
\left(
\bm{m}_{i\mid c}(t),
\bm{K}_{i\mid c}(t)
\right)\ ,
\label{eq:13}
\end{eqnarray}
where the posterior mean and covariance functions are given by
\begin{eqnarray}
\hspace{-3.5mm}\bm{m}_{i\mid c}(t)
&=&
\bm{m}_i(t\mid\bm{x},\bm{x}')
+
\bm{K}_{ic}(t\mid\bm{x},\bm{x}')
\bm{K}_{cc}(t\mid\bm{x},\bm{x}')^{-1}
\left(
\bm{x}_c(t\mid\bm{x},\bm{x}')
-
\bm{m}_c(t\mid\bm{x},\bm{x}')
\right)\ ,
\nonumber\\
\hspace{-18mm}\bm{K}_{i\mid c}(t)
&=&
\bm{K}_{ii}(t\mid\bm{x},\bm{x}')
-
\bm{K}_{ic}(t\mid\bm{x},\bm{x}')
\bm{K}_{cc}(t\mid\bm{x},\bm{x}')^{-1}
\bm{K}_{ci}(t\mid\bm{x},\bm{x}')\ .
\label{eq:14}
\end{eqnarray}
As such, $\bm{m}_{i\mid c}(t) = \mathbb{E}[\bm{x}_i(t\mid \bm{x},\bm{x}')]$ and we can use this posterior mean as the correlated multi-physics interpolation or structure proposer that supplies the learning signal for the flow matching loss in Eq.~\eqref{eq:9}.~This means that we optimize the flow matching loss using the average interpolation $\mathbb{E}[\bm{x}_i(t\mid \bm{x},\bm{x}') ] = \bm{m}_{i\mid c}(t)$ rather than the sampled interpolation $\bm{x}_i(t\mid \bm{x},\bm{x}')$.~Consequently, the resulting loss is given by:
\begin{eqnarray}
\hspace{-16mm}\theta &=& \arg\min_\theta \frac{1}{m}\sum_{i=1}^m\mathbb{E}_{s,t} \mathbb{E}_{(\bm{x}, \bm{x}')} \mathbb{E}_{\bm{x}_i(t)}\Big\|\bm{u}_\theta^i\big(\mathbb{E}[\bm{x}(t)], t\big) \ \ -\ \  \mathrm{d}\mathbb{E}[\bm{x}_i(t\mid \bm{x},\bm{x}')]/\mathrm{d}t\Big\|^2 \ ,\label{eq:15}
\end{eqnarray}
Although this changes the form of the loss in Eq.~\eqref{eq:9}, it does not change its theoretical optimizer -- see Theorem~\ref{thm:1} below.~An important advantage is that although $\bm{x}_i(t\mid \bm{x},\bm{x}')$ is stochastic and may not be differentiable, its posterior mean $\bm{m}_{i\mid c}(t)$ has a closed form in Eq.~\eqref{eq:14} allowing us to compute the corresponding instantaneous velocity $\mathrm{d}\mathbb{E}[\bm{x}_i(t\mid \bm{x},\bm{x}')]/\mathrm{d}t = \mathrm{d}\bm{m}_{i\mid c}(t)/\mathrm{d}t$ which is required for computing the flow matching loss in Eq.~\eqref{eq:9}.
\begin{theorem}\label{thm:1}
The theoretical optimizers of the conditional mean flow matching losses in Eq.~\eqref{eq:9} and Eq.~\eqref{eq:15} are the same.~A detailed proof is provided in Appendix~\ref{appendix:proof_thm_1}.
\end{theorem}

The key advantage of GP-based interpolation is its ability to mitigate the structural bias introduced by independent linear interpolation.
In particular, linear interpolation determines each process's intermediate state solely from its own endpoints, without explicitly accounting for how other processes evolve along the transport path.
This can be restrictive in coupled multi-physics systems, where the intermediate evolution of one process depends on the states of others.

In contrast, the GP posterior mean $\bm{m}_{i\mid c}(t)$ explicitly conditions the interpolation of process $i$ on the intermediate state of process $c$ through the cross-covariance operators $\bm{K}_{ic}$ and $\bm{K}_{ci}$.
Consequently, the proposed intermediate states can reflect cross-process dependencies beyond those implicitly encoded by the endpoint pairs.~In practice, we can initialize the interpolation for one process $\bm{x}_c(t\mid\bm{x},\bm{x}')$ using standard linear interpolation between its endpoints and subsequently generate correlated interpolations for the remaining processes using the posterior GP marginals in Eq.~\eqref{eq:13}.~This provides a practical mechanism for constructing coupled intermediate states while incorporating cross-physics correlations through the GP covariance operators. Moreover, when conditioned on the endpoint observations, the GP-based interpolations preserve both endpoint conditions, as theoretically established in Proposition~\ref{prop:gp_endpoint}.


For practical implementation, we use zero-mean Gaussian processes~\citep{Rasmussen06}, a standard choice such that the prior mean is zero: $\bm{m}_{i}(t \mid \bm{x}, \bm{x}') =0$.~The covariance operators $\bm{K}_{ic}(t,t'\mid\bm{x},\bm{x}')$ are defined using smooth kernels, such as the squared-exponential kernel~\citep{Rasmussen06}, which encourage temporally smooth interpolations while capturing correlations among dependent physical processes.

\subsection{Coupled Mean-Flow Matching: Algorithms and Guarantees}
\label{subsec:algo}
{\bf A.~Workflow.}~The two key components developed in Section~\ref{subsec:identity} and Section~\ref{subsec:gp} are integrated into the proposed \textsc{CoupledFlow} algorithm for learning a multi-physics neural operator (see Alg.~\ref{alg:training}) that enables one-step prediction at inference time (see Alg.~\ref{alg:sampling}).

\begin{algorithm}[h!]
    \caption{Constructing Multi-Physics Neural Operator}
    \label{alg:training}
    \begin{algorithmic}[1]
        \STATE \textbf{input:} simulation dataset $D$, number of training epochs $n$, prediction time $\tau > 0$
        \STATE \textbf{output:} optimized coupled mean flow parameter $\theta$.
        \FOR{$i = 1$ to $n$}
            \STATE  sample $s \in \mathcal{U}(0,\tau)$ and $t \in \mathcal{U}(s, \tau)$
            \STATE sample $(\bm{x}, \bm{x}') \sim D$
            \STATE sample randomly $c \in [m]$ and construct $\bm{x}_c(t\mid \bm{x},\bm{x}') =  (1-t/\tau)\bm{x}_c + (t/\tau)\bm{x}'_c$
            \STATE compute $\mathbb{E}[\bm{x}_i(t\mid \bm{x},\bm{x}')]$ for each $i \in [m], i \ne c$ via Eq.~\eqref{eq:14}
            \STATE compute the learning signal $\mathrm{d}\mathbb{E}[\bm{x}_i(t\mid \bm{x},\bm{x}')]/\mathrm{d}t$ for Eq.~\eqref{eq:15} 
            \STATE update $\theta$ via running Adam optimizer~\citep{kingma2015adam} on Eq.~\eqref{eq:15}
        \ENDFOR
        \STATE \textbf{return} the optimized coupled mean flow parameterization $\theta$
    \end{algorithmic}
\end{algorithm}

\begin{algorithm}[h!]
    \caption{One-Step Prediction with Coupled Mean Flow}
    \label{alg:sampling}
    \begin{algorithmic}[1]
        \STATE \textbf{input:} initial multi-physics state $\bm{x}(0)=(\bm{x}_1(0),\ldots,\bm{x}_m(0))$
        \STATE \textbf{output:} predicted multi-physics solution $\hat{\bm{x}}(\tau) =(\hat{\bm{x}}_1(\tau),\ldots,\hat{\bm{x}}_m(\tau))$
        \FOR{$i = 1$ to $m$}
            \STATE $\hat{\bm{x}}_i(\tau) \leftarrow \bm{x}_i(0) \ \ +\ \  \tau \,\bar{\bm{u}}_\theta^i(\bm{x}(0),0,\tau)$
        \ENDFOR
        \STATE \textbf{return} $\hat{\bm{x}}(\tau)=(\hat{\bm{x}}_1(\tau),\ldots,\hat{\bm{x}}_m(\tau))$
    \end{algorithmic}
\end{algorithm}


{\bf B.~Performance Guarantee.}
~We establish a theoretical guarantee relating the proposed algorithm's training objective to its test-time prediction error.
~Given a test-time distribution $\bm{x}_i(0) \sim \pi_s^i(\bm{x})$, let $\pi_\tau^i(\bm{x})$ denote the marginal distribution over the corresponding (oracle) output $\bm{x}(\tau)$.~Similarly, let $\bar{\pi}^i_\tau(\bm{x}')$ denote the marginal distribution over the predicted outputs of our multi-physics neural operator \textsc{CoupledFlow} when testing inputs are sampled from $\pi_s^i(\bm{x})$.

We measure test-time prediction error $\textsc{err}(\theta)$ using the Wasserstein-2 distance (on $\ell_2$ cost metric) between $\pi_\tau^{i}$ and $\bar{\pi}_\tau^i$, averaged across all physical processes: $\textsc{err}(\theta) = (1/m)\sum_{i=1}^m W_2(\pi_\tau^i, \bar{\pi}_\tau^i)$.~The following theorem establishes an upper bound on this test-time prediction error in terms of the coupled mean-flow matching loss in Eq.~\eqref{eq:15}.

\begin{theorem}\label{thm:2}

Let $\ell_r$ denote the Lipschitz constant of the instantaneous multi-physics velocity $\bm{u}_\theta(\bm{x}(r), r) = (\bm{u}_\theta^1(\bm{x}(r), r), \ldots, \bm{u}_\theta^m(\bm{x}(r), r))$.
Let $\kappa \triangleq \exp(\int_0^\tau \ell_r\mathrm{d}r)$, it follows that
\begin{eqnarray}
\textsc{err}(\theta) &\leq& \kappa \tau \sqrt{L(\theta)} ,
\end{eqnarray}
where $L(\theta)$ denotes the coupled mean flow matching loss in Eq.~\eqref{eq:15} which is restated below:
\begin{eqnarray}\label{eq: L_theta}
\hspace{-23.5mm}L(\theta) &\triangleq& \frac{1}{m}\sum_{i=1}^m\mathbb{E}_{s,t} \mathbb{E}_{(\bm{x}, \bm{x}')} \mathbb{E}_{\bm{x}_i(t)}\Big\|\bm{u}_\theta^i\big(\mathbb{E}[\bm{x}(t)], t\big) \ \ -\ \  \mathbb{E}[\mathrm{d}\bm{x}_i(t\mid \bm{x},\bm{x}')/\mathrm{d}t]\Big\|^2 \ .
\end{eqnarray}
A detailed proof of this theorem is provided in Appendix~\ref{appendix:proof_theorem_2}.~Here, $\tau$ denotes the prediction time.  
\end{theorem}
This result implies that the generalization error can be small when (1) $\theta$ is well optimized, resulting in a small training loss $L(\theta)$; and (2) the Lipschitz constant $\ell_r$ of the instantaneous multi-physics velocity $u_\theta(\bm{x}(r), r)$ is small.~This reflects the interplay between the algorithm and the intrinsic complexity of the physics system.~For PDE systems with rapidly varying dynamics, the Lipschitz constant $\ell_r$ can be large due to the sensitivity of the instantaneous velocity to changes in the system state.
~As a consequence, learning accurate neural operators for such systems can be more challenging, and the resulting bound on test-time prediction error may be less informative.~The bound becomes less informative for longer prediction horizons $\tau$, as expected.~The tightness of this error bound is also validated empirically in Appendix~\ref{appendix:Experiment_detailed_results} and Figure~\ref{fig:theorem2_bound}.

\section{Empirical Evaluation and Analysis} \label{sec:experiments} 

This section presents empirical results comparing \textsc{CoupledFlow} and SOTA baselines. The official code repository is provided at ~\url{https://github.com/khiembk/CoupledFlow}.

\subsection{Experimental Setup}
\textbf{Benchmarks}:  Following prior work \citep{sun2025compolunifiedneuraloperator,xiao2023coupled}, we evaluate \textsc{Coupled-Flow} on five coupled multi-physics PDE benchmarks: 1-D Lotka-Volterra equations \citep{Murray2002} (2 processes), 1-D Belousov-Zhabotinsky equations \citep{Petrov1993} (3 processes), 2-D Gray-
Scott equations \citep{Pearson_1993} (2 processes), 2-D multiphase flow of oil and water in geological
storage \citep{bear2010modeling} \citep{hashemi2021pore} \cite{aboukassem2013petroleum} (2 processes), and a 2-D Thermal-Hydro-Mechanical problem \citep{Gao2020} (5 processes). We also evaluate on two PDE benchmarks: Diffusion Reaction-2D (2 processes) from PDEBench \citep{takamoto2024pdebenchextensivebenchmarkscientific} and Navier-Stokes-2D (3 processes) from PDEArena~\citep{gupta2022towards}.

\textbf{Baselines}: We compare \textsc{CoupledFlow} against state-of-the-art neural operators and generative PDE solvers: COMPOL \citep{sun2025compolunifiedneuraloperator}, FNO \citep{li2021fourier}, CViT~\citep{wang2025cvitcontinuousvisiontransformer}, UFNO \citep{wen2022ufnoenhancedfourier}, DiffusionPDE~\citep{huang2024diffusionpde}, Transolver \citep{wu2024transolverfasttransformersolver}, and CMWNO \citep{xiao2023coupled}. COMPOL and CMWNO are specifically designed for coupled PDEs, while the remaining baselines are not explicitly designed for multi-physics modeling. 


\textbf{Evaluation Methodology}: We evaluate prediction accuracy using the relative $L_2$ error between predicted and ground-truth PDE solutions, defined as
\begin{equation}
\mathrm{Error}_{L_2} = \frac{\|\hat{\bm{z}}-\bm{z}\|_2}{\|\bm{z}\|_2}\ , 
\end{equation}
where $\hat{\bm{z}}$ and $\bm{z}$ denote the predicted and ground-truth solution vectors obtained by concatenating the outputs of all physical processes following prior work~\citep{sun2025compolunifiedneuraloperator, xiao2023coupled}. We consider two training-data regimes with 512 and 1024 samples to assess performance under different levels of data availability. All experiments are repeated over five independent runs, and we report the mean prediction error, with standard deviations in Appendix~\ref{appendix:Experiment_detailed_results}. 
We also evaluate long-horizon rollout performance, robustness to distribution shifts, and computational efficiency. Experimental configurations and hyperparameters are provided in Appendix~\ref{appdix:detailed_exp_setup}.

\subsection{Main Results}

{\bf Predictive Performance.} The results in Table~\ref{tab:main_results} and Table~\ref{tab:nrmse_main_std} show the predictive performance of \textsc{Coupled-Flow} against baseline neural operators (COMPOL, FNO, UFNO, Transolver, CMWNO) and recent approaches based on vision transformers, CViT~\citep{wang2025cvitcontinuousvisiontransformer} and diffusion models, DiffusionPDE~\citep{huang2024diffusionpde}. We evaluate all methods with 512 and 1024 training samples across five datasets (Lotka-
Volterra, Belousov-Zhabotinsky, Gray- Scott, multiphase flow, and Thermal-Hydro-Mechanical). We make the following observations. 

{\bf 1.}~Across both training-data regimes, \textsc{CoupledFlow} with the GP interpolant (\textsc{coupled-flow}-GP) achieves the best predictive performance on 7 out of 10 tasks and the second-best performance on the remaining 3 tasks. 

{\bf 2.}~\textsc{CoupledFlow} with the linear interpolant (\textsc{coupled-flow}-Linear) also performs competitively, achieving second-best performance on 7 out of 10 tasks. The improvements are particularly pronounced on benchmarks with strong cross-process interactions.

To evaluate the generality of \textsc{CoupledFlow} beyond the coupled PDE benchmarks considered in prior coupled-operator work, we conduct additional experiments on two standard benchmarks: Reaction-Diffusion-2D from PDEBench \citep{takamoto2024pdebenchextensivebenchmarkscientific} and Navier-Stokes-2D from PDEArena~\citep{gupta2022towards}. Table~\ref{tab:rd_ns} reports the relative $L_2$ prediction errors. We include CViT-B \citep{wang2025cvitcontinuousvisiontransformer} as a strong baseline, given its recent state-of-the-art performance across several operator-learning benchmarks. 

{\bf 3)} On Reaction-Diffusion-2D, \textsc{CoupledFlow}-GP achieves performance close to CViT-B, the strongest baseline, while substantially outperforming the other neural-operator and coupled-operator baselines. On Navier-Stokes-2D, \textsc{CoupledFlow}-GP achieves the best predictive performance.

\begin{table*}[t]
\centering
\normalsize
\renewcommand{\arraystretch}{0.75}
\setlength{\tabcolsep}{4.2pt}
\begin{tabular}{@{}lrrrrr@{}}
\toprule
\textbf{Method} 
& \textbf{Lotka-V.} 
& \textbf{Belousov-Z.} 
& \textbf{Gray-Scott} 
& \textbf{Multiphase} 
& \textbf{THM} \\
\midrule

$\text{FNO}_{c}$ 
& 9.55 & 0.16 & 0.55 & 2.91 & 0.11 \\

\midrule

CMWNO 
& 6.65 & 0.14 & 0.43 & 1.93 & 0.09 \\

\midrule

$\text{UFNO}_{c}$ 
& 8.73 & 0.17 & 0.36 & 3.87 & 0.08 \\

\midrule

$\text{Transolver}_{c}$ 
& 17.67 & 1.88 & 0.49 & 0.88 & 0.16 \\

\midrule

$\text{DiffusionPDE}_{c}$ 
& 3.10 & 0.53 & 0.21 & 2.45 & 0.16 \\

\midrule

$\text{MeanFlow}_{c}$ 
& 13.21 & 0.17 & 0.29 & 3.33 & 0.13 \\

\midrule

CViT-B 
& \textbf{0.77} & 0.31 & 0.74 & 1.39 & 0.51 \\

\midrule

COMPOL-RNN 
& 8.02 & 0.15 & 0.50 & 6.44 & 0.09 \\

\midrule

COMPOL-ATN 
& 8.58 & 0.15 & 0.48 & 6.76 & 0.10 \\

\midrule

\textsc{CoupledFlow}-Linear 
& 1.29  
& \textcolor{blue}{\textbf{0.09}}  
& \textcolor{blue}{\textbf{0.18}}  
& \textbf{0.09} 
& \textcolor{blue}{\textbf{0.08}} \\

\midrule

\textsc{CoupledFlow}-GP 
& \textcolor{blue}{\textbf{1.21}} 
& \textbf{0.06} 
& \textbf{0.15} 
& \textcolor{blue}{\textbf{0.11}} 
& \textbf{0.05} \\

\bottomrule
\end{tabular}

\caption{Comparison of relative $L_2$ prediction errors $(\times 10^{-2})$ across five multi-physics benchmark datasets with 1024 training samples. \textbf{Bold} and \textcolor{blue}{\textbf{blue}} values indicate the best and second-best performance, respectively. Detailed results with standard deviations are provided in Table~\ref{tab:nrmse_main_std}.}
\label{tab:main_results}
\vspace{-5mm}
\end{table*}

\subsection{Ablation Studies}

\textbf{Rollout Prediction}. We further evaluate whether the learned one-step operator remains stable when recursively applied
for long-horizon prediction. Tables~\ref{tab:rollout_gs_std} and~\ref{tab:rollout_lv_std} report rollout errors on Lotka--Volterra and Gray--Scott using $512$ training samples and $256$ independent test samples, over 10 rollout steps. \textsc{CoupledFlow} consistently achieves
the best or second-best performance across all rollout horizons, indicating that the proposed transport formulation not only improves single-step prediction but also supports stable multi-step forecasting.

\textbf{Component-wise Contribution.} Table~\ref{tab:separate_contribution} evaluates the contribution of the coupled transport formulation relative to the vanilla MeanFlow baseline. {\bf (1)} Both \textsc{CoupledFlow}-GP and \textsc{CoupledFlow}-Linear substantially outperform $\text{MeanFlow}_{c}$ across all data regimes, demonstrating that conditioning process-specific average velocities on the full coupled state is important for multi-physics prediction. {\bf (2)} The GP variant further improves over the linear variant.
This suggests that GP-based coupled paths provide more informative intermediate supervision than independent linear interpolation.

\begin{wraptable}{r}{0.48\textwidth}
\centering
\resizebox{\linewidth}{!}{%
\begin{tabular}{lcccc}
\toprule
\multirow{2}{*}{\textbf{Method}} 
& \multicolumn{2}{c}{\textbf{Lotka-V.}} 
& \multicolumn{2}{c}{\textbf{Belousov-Z.}} \\
\cmidrule(lr){2-3} \cmidrule(lr){4-5}
& \textbf{512} & \textbf{1024} 
& \textbf{512} & \textbf{1024}  \\
\midrule
$\text{MeanFlow}_{c}$    
& 16.52 & 13.21 & 0.17 & 0.17 \\
\midrule
\textsc{CoupledFlow}-GP     
& \textbf{2.16} & \textbf{1.21} & \textbf{0.11} & \textbf{0.06} \\
\textsc{CoupledFlow}-Linear 
& \textcolor{blue}{\textbf{2.34}} 
& \textcolor{blue}{\textbf{1.29}} 
& \textcolor{blue}{\textbf{0.15}} 
& \textcolor{blue}{\textbf{0.09}} \\
\bottomrule
\end{tabular}%
}
\vspace{-1mm}
\caption{Comparison of relative $L_2$ errors $(\times 10^{-2})$ on Lotka--Volterra and Belousov--Zhabotinsky. Detailed results with standard deviations are provided in Table~\ref{tab:nrmse_main_std} of Appendix~\ref{appendix:Experiment_detailed_results}.}
\label{tab:separate_contribution}
\end{wraptable}

\begin{wraptable}{t}{0.48\columnwidth}
\vspace{-47mm}
\centering
\normalsize
\renewcommand{\arraystretch}{1}
\setlength{\tabcolsep}{2pt}
\begin{tabular}{@{}lrr@{}}
\toprule
\textbf{Method} 
& \textbf{RD-2D} 
& \textbf{NS-2D} \\
\midrule
$\text{MeanFlow}_{c}$ & 9.51 & 16.13 \\
$\text{FNO}_{c}$ & 12.34 & 9.12 \\
$\text{UFNO}_{c}$ & 12.33 & 9.47 \\
COMPOL-RNN & 8.28 & 8.25 \\
COMPOL-ATN & 8.27 & 8.21 \\
CMWNO & 8.45 & 8.42 \\
DiffusionPDE & 12.65 & 8.84 \\
CViT-B & \textbf{1.58} & \textcolor{blue}{\textbf{3.95}} \\
\midrule
\textsc{CoupledFlow}-GP 
& \textcolor{blue}{\textbf{1.63}} 
& \textbf{3.45} \\
\textsc{CoupledFlow}-Linear 
& 1.74 
& 3.98 \\
\bottomrule
\end{tabular}
\caption{Relative $L_2$ prediction error $(\times 10^{-2})$ on Reaction-Diffusion-2D (RD-2D) and Navier-Stokes-2D (NS-2D). Detailed results with standard deviations are provided in Appendix~\ref{appendix:Experiment_detailed_results}.}
\label{tab:rd_ns}
\end{wraptable}



\textbf{Sensitivity and Robustness Analysis. } We examine the sensitivity of \textsc{CoupledFlow}-GP to the GP kernel and length-scale. 
{\bf (1)} As shown in Table~\ref{tab:lengthscale_sens_std}, performance remains stable across a broad range of $\ell$, indicating that the method does not require precise length-scale tuning. 
{\bf (2)} Table~\ref{tab:kernel_sensitivity} shows that several standard kernels, including squared-exponential, Matérn-$3/2$, Matérn-$5/2$, and rational quadratic, perform close to or better than the linear interpolation baseline. 
{\bf (3)} Since different kernels perform best on different datasets, the gains appear to come from correlated GP-based intermediate paths rather than reliance on a specific covariance function. 
{\bf (4)} Additional OOD coefficient-shift results are provided in Table~\ref{tab:ood_coeff} of Appendix~\ref{appendix:Experiment_detailed_results}. \textsc{CoupledFlow}-GP consistently achieves the best or second-best generalization performance, highlighting the robustness of our method. Table~\ref{tab:mpf_ood} also reports experiments under varying initial and boundary conditions on MPF with matched model capacities, where \textsc{CoupledFlow}-GP outperforms DiffusionPDE in both ID and OOD settings. {\bf (5)} Results for boundary and initial-condition
prediction errors are provided in Table~\ref{tab:additional_metrics}
of Appendix~\ref{appendix:Experiment_detailed_results}.
\textsc{CoupledFlow} achieves the lowest overall and boundary
errors on both MPF and THM benchmarks, further demonstrating
its effectiveness in capturing coupled PDE solution behavior. 

\begin{table}[t]
\centering
\small
\begin{tabular}{lrrrr}
\toprule
\textbf{Method} 
& \textbf{Lotka-V.} 
& \textbf{Belousov-Z.} 
& \textbf{Gray-Scott} 
& \textbf{Multiphase} 
 \\
\midrule
$\text{FNO}_{c}$         & 0.70 & 10.15 & 0.20 & 2.04 \\
$\text{UFNO}_{c}$        & 0.65 & 10.04 & 0.25 & 5.27 \\
$\text{CMWNO}_{c}$       & 0.62 & 7.11 & 0.22 & 2.36 \\
$\text{COMPOL-ATN}_{c}$  & 0.48 & 8.74 & 0.18 & 2.67 \\
$\text{COMPOL-RNN}_{c}$  & 0.52 & 8.52 & 0.19 & 1.87 \\
CViT        & 0.42 & 3.70 & 1.13 & 3.26 \\
\midrule
\textsc{CoupledFlow}-Linear
            & 0.23 & 2.30 & 0.15 & 0.90 \\
\textsc{CoupledFlow}-GP
            & \textbf{0.19}
            & \textbf{2.10}
            & \textbf{0.11}
            & \textbf{0.61} \\
\bottomrule
\end{tabular}
\vspace{2mm}
\caption{Comparison of relative $L_2$ prediction errors $(\times 10^{-2})$
with approximately matched parameter counts.
The parameter budgets are 10M, 9M, 13M, and 10M for
Lotka--Volterra, Belousov--Zhabotinsky, Gray--Scott, and Multiphase, respectively.
The best results are highlighted in bold.}
\label{tab:scale_analysis}
\end{table}

\textbf{Model Scale and Computational Efficiency.}~To investigate whether the performance gains of \textsc{CoupledFlow} are attributable to larger model capacity, we conduct 
experiments with approximately matched parameter counts across
methods. Specifically, we consider parameter budgets of
10M, 9M, 13M, and 10M for Lotka--Volterra, Belousov--Zhabotinsky, Gray--Scott, and Multiphase, respectively. As shown in Table~\ref{tab:scale_analysis},
\textsc{CoupledFlow}-GP consistently achieves the lowest prediction
error across all four benchmarks, reducing the error by
38.9\%--67.4\% relative to the strongest competing baseline
on each dataset. In terms of computational efficiency, our method introduces a
small inference-time overhead compared to FNO and CViT,
averaging approximately 3.8 ms and 3.7 ms, respectively,
across the five benchmarks reported in Table~\ref{tab:infer_time}.
Importantly, the GP-based interpolations are used only
during training, while inference remains a one-step operation.
Overall, the substantial improvements in prediction accuracy
come at a modest additional inference cost, indicating a
favorable accuracy--efficiency trade-off.

{\bf Runtime/Speed Results.} Due to space constraints, we report these results in Table~\ref{tab:infer_time} of Appendix. \textsc{CoupledFlow} achieves substantially faster inference than iterative generative baselines such as DiffusionPDE and is consistently faster than CMWNO. Its inference runtime is also competitive with standard neural operators and transformer-based methods.\vspace{-2mm}

\section{Conclusion} \label{sec:conclusion}

We introduced \textsc{CoupledFlow}, a one-step flow-based neural operator for coupled multi-physics PDE systems. Unlike standard endpoint-regression neural operators, our method learns transport dynamics through process-specific average velocity fields conditioned on the full coupled state, capturing cross-process interactions during training. To provide informative path-level supervision, we constructed Gaussian process-based correlated interpolants that capture cross-process dependencies more effectively than independent linear interpolants. More broadly, our results show that intermediate transport paths can guide the learning of cross-physics dependencies without requiring their explicit simulation at inference time. Our theoretical analysis established that the prediction error of the learned operator can be controlled by the coupled mean-flow matching loss and the smoothness of the induced transport. Empirically, \textsc{CoupledFlow} outperformed state-of-the-art baselines across diverse coupled PDE benchmarks.

\section*{Acknowledgement}
This research is supported by the National Science Foundation CAREER Award IIS-2544071. Phi Le Nguyen is supported by the Vietnam National Foundation for Science and Technology Development (NAFOSTED) under project code 24/2025/HĐ-TN.01.
\bibliographystyle{plainnat}
\bibliography{references}

\newpage
\appendix
\section{Proof of Theorem~\ref{thm:1}}
\subsection{Endpoint Consistency of GP interpolation}
\begin{proposition}
\label{prop:gp_endpoint}
Let $\bm{x}(t\mid\bm{x},\bm{x}')$ be a multi-output Gaussian
process conditioned on the endpoint observations
$\bm{x}(0)=\bm{x}$ and $\bm{x}(1)=\bm{x}'$.
Then the GP posterior mean satisfies
\begin{eqnarray}
     \bm{m}_{i\mid c}(0)=\bm{x}_i,
    \qquad
    \bm{m}_{i\mid c}(1)=\bm{x}'_i, \nonumber
\end{eqnarray}

for every physical process $i$, where
$\bm{m}_{i\mid c}(t)$ denotes the posterior mean of process $i$
conditioned on process $c$.
\end{proposition}

\begin{proof}
We consider a zero-mean multi-output Gaussian process with
covariance kernel $\bm{K}_{ic}(t,s \mid \bm{x}, \bm{x}')$.
Let the endpoint observations be
\begin{eqnarray}
     \bm{y}_i =
    \begin{bmatrix}
        \bm{x}_i\\
        \bm{x}_i'
    \end{bmatrix},
    \qquad
    \mathcal{T}=\{0,1\}. \nonumber
\end{eqnarray}

Conditioning the GP on these observations yields the
posterior mean
\begin{equation}
    \bm{m}_{i \mid c}(t)
    =
    \bm{K}_{ic}(t,\mathcal{T} \mid \bm{x}, \bm{x}')
    \bm{K}_{cc}(\mathcal{T},\mathcal{T} \mid \bm{x}, \bm{x}')^{-1}
    \bm{y}_i,
\end{equation}
and posterior covariance
\begin{equation}
    \bm{K}_{i \mid c}(t)
    =
    \bm{K}_{ic}(t \mid \bm{x}, \bm{x}')
    -
    \bm{K}_{ic}(t,\mathcal{T} \mid \bm{x}, \bm{x}')
    \bm{K}_{cc}(\mathcal{T},\mathcal{T} \mid \bm{x}, \bm{x}')^{-1}
    \bm{K}_{ci}(\mathcal{T},s \mid \bm{x}, \bm{x}').
\end{equation}

At $t=0$, since $\bm{K}_{ic}(0,\mathcal{T} \mid \bm{x}, \bm{x}')$ corresponds to
the first block row of $\bm{K}_{cc}(\mathcal{T},\mathcal{T} \mid \bm{x}, \bm{x}')$,
we obtain
\begin{eqnarray}
      \bm{m}_{i \mid c}(0)=
    \bm{K}_{ic}(0,\mathcal{T} \mid \bm{x}, \bm{x}')
    \bm{K}_{cc}(\mathcal{T},\mathcal{T} \mid \bm{x}, \bm{x}')^{-1}
    \bm{y}_i = \bm{x}_i
\end{eqnarray}

Similarly, at $t=1$,
\begin{eqnarray}
    \bm{m}_{i \mid c}(1)=
    \bm{K}_{ic}(1,\mathcal{T} \mid \bm{x}, \bm{x}')
    \bm{K}_{cc}(\mathcal{T},\mathcal{T} \mid \bm{x}, \bm{x}')^{-1}
    \bm{y}_i = \bm{x}'_i
\end{eqnarray}

Furthermore, the posterior covariance vanishes at
the observed endpoints:
\begin{equation}
  \bm{K}_{i \mid c}(0) = \bm{K}_{i \mid c}(1) = 0
\end{equation}

Therefore, conditioning on an additional physical process
$c$ cannot alter the deterministic endpoint values
of process $i$.
This establishes that the GP-based correlated interpolants
preserve both prescribed endpoints.
\end{proof}
\subsection{Proof of Theorem~\ref{thm:1}}
\label{appendix:proof_thm_1}
\begin{theorem*}
     Theoretic optimizers of the conditional mean flow matching losses in Eq.~\eqref{eq:9} and Eq.~\eqref{eq:15} are the same.
\end{theorem*}
\begin{proof}
    We can express the objective function in Eq.~\eqref{eq:9} as 
    \begin{eqnarray}
        L (\theta) &=& \frac{1}{m} \sum_{i=1}^{m} \mathbb{E}_{(s, t)} \mathbb{E}_{(\bm{x}, \bm{x}')} \mathbb{E}_{\bm{x}_i(t)}\Big\| \bm{u}^{i}_{\theta}\big(\mathbb{E}[\bm{x}(t)], t \big) - d \bm{x}_i(t \mid \bm{x}, \bm{x}')/dt\Big\|^2 \\ \nonumber
        &=& \frac{1}{m} \sum_{i=1}^{m} \mathbb{E}_{(s, t)} \mathbb{E}_{(\bm{x}, \bm{x}')} \mathbb{E}_{\bm{x}_i(t)}\Big\| \bm{u}^{i}_{\theta}\big(\mathbb{E}[\bm{x}(t)], t \big) - d\mathbb{E}[\bm{x}_i(t \mid \bm{x}, \bm{x}')]/dt\Big\|^2  \\ \nonumber
        &+& \frac{1}{m} \sum_{i=1}^{m} \mathbb{E}_{(s, t)} \mathbb{E}_{(\bm{x}, \bm{x}')} \mathbb{E}_{\bm{x}_i(t)}\Big\|  d\mathbb{E}[ \bm{x}_i(t \mid \bm{x}, \bm{x}')]/dt - d \bm{x}_i(t \mid \bm{x}, \bm{x}')/dt\Big\|^2  \\ \nonumber
        &+& \frac{2}{m} \sum_{i=1}^{m} \mathbb{E}_{(s, t)} \mathbb{E}_{(\bm{x}, \bm{x}')} \mathbb{E}_{\bm{x}_i(t)} \langle \bm{u}^{i}_{\theta}\big(\mathbb{E}[\bm{x}(t)], t \big) - d\mathbb{E}[ \bm{x}_i(t \mid \bm{x}, \bm{x}')]/dt ,  d\mathbb{E}[ \bm{x}_i(t \mid \bm{x}, \bm{x}')]/dt - d \bm{x}_i(t \mid \bm{x}, \bm{x}')/dt \rangle \\ \nonumber \end{eqnarray}
    Because  $\mathbb{E}_{\bm{x}_i(t)} [d\mathbb{E}[ \bm{x}_i(t \mid \bm{x}, \bm{x}')]/dt - d \bm{x}_i(t \mid \bm{x}, \bm{x}')/dt] = 0$, we have 
    \begin{eqnarray}
     L(\theta)   &=& \frac{1}{m} \sum_{i=1}^{m} \mathbb{E}_{(s, t)} \mathbb{E}_{(\bm{x}, \bm{x}')} \mathbb{E}_{\bm{x}_i(t)}\Big\| \bm{u}^{i}_{\theta}\big(\mathbb{E}[\bm{x}(t)], t \big) - d\mathbb{E}[\bm{x}_i(t \mid \bm{x}, \bm{x}')]/dt\Big\|^2  \\ \nonumber
        &+& \frac{1}{m} \sum_{i=1}^{m} \mathbb{E}_{(s, t)} \mathbb{E}_{(\bm{x}, \bm{x}')} \mathbb{E}_{\bm{x}_i(t)}\Big\|  d\mathbb{E}[ \bm{x}_i(t \mid \bm{x}, \bm{x}')]/dt - d \bm{x}_i(t \mid \bm{x}, \bm{x}')/dt\Big\|^2  \\ \nonumber 
        &=& \frac{1}{m} \sum_{i=1}^{m} \mathbb{E}_{(s, t)} \mathbb{E}_{(\bm{x}, \bm{x}')} \mathbb{E}_{\bm{x}_i(t)} \Big\| \bm{u}^{i}_{\theta}\big(\mathbb{E}[\bm{x}(t)], t \big) - d\mathbb{E}[\bm{x}_i(t \mid \bm{x}, \bm{x}')/dt]\Big\|^2 + \text{const}
   \end{eqnarray}
    Here, the second-to-last equality follows from the interchangeability between the expectation $\mathbb{E}$ and the derivative $d/dt$ of $\bm{x}_i(t \mid \bm{x}, \bm{x}')$ when $\bm{x}_i(t \mid \bm{x}, \bm{x}')$ follows the law of GPs with a squared exponential kernel, which we used in our experiments. This shows that Eq.~\eqref{eq:9} and~\eqref{eq:15}  only differ by a constant.
\end{proof}

\section{Proof of the Theorem~\ref{thm:2}}
\label{appendix:proof_theorem_2}

\subsection{Key lemmas}
We need the following lemmas to prove Theorem \ref{thm:2}.

\begin{lemma}\label{lem:gronwall}
Let \(\ell_r\) denote the Lipschitz constant of
\(\bm{u}_\theta(\bm{x}(r), r)\) with respect to its spatial argument, and define
\begin{eqnarray}
    \kappa
&\triangleq&
\exp\!\left(
\int_0^\tau \ell_r\,dr
\right).
\end{eqnarray}

Then the operator norm of the Jacobian of the state
\(\bm{x}(\tau)\) with respect to the initial condition
\(\bm{x}(0)\) satisfies
\begin{eqnarray}
\|\nabla_{\bm{x}(0)} \bm{x}(\tau)\|_{\mathrm{op}}
&\leq&
\kappa.
\end{eqnarray}
\end{lemma}

\begin{proof}
Let \(\bm{x}(r)\) be the solution to the ODE
\begin{eqnarray}
    \frac{d}{dr}\bm{x}(r)
&=&
\bm{u}_\theta(\bm{x}(r), r),
\qquad
\bm{x}(0)=\bm{x}.
\end{eqnarray}

Define the Jacobian matrix
\begin{eqnarray}
J(r)
&\triangleq&
\nabla_{\bm{x}(0)} \bm{x}(r).
\end{eqnarray}

Differentiating the dynamics with respect to the initial condition
\(\bm{x}(0)\) yields the variational equation
\begin{eqnarray}
\frac{d}{dr}J(r)
&=&
\nabla_{\bm{x}}
\bm{u}_\theta(\bm{x}(r), r)\, J(r),
\qquad
J(0)=I.
\end{eqnarray}

Taking operator norms and using submultiplicativity,
\begin{eqnarray}
\frac{d}{dr}\|J(r)\|_{\mathrm{op}}
&\leq&
\left\|
\nabla_{\bm{x}}
\bm{u}_\theta(\bm{x}(r), r)
\right\|_{\mathrm{op}}
\,
\|J(r)\|_{\mathrm{op}}.
\end{eqnarray}

Since \(\bm{u}_\theta(\cdot,r)\) is \(\ell_r\)-Lipschitz,
\begin{eqnarray}
\left\|
\nabla_{\bm{x}}
\bm{u}_\theta(\bm{x}(r), r)
\right\|_{\mathrm{op}}
&\leq&
\ell_r.
\end{eqnarray}

Therefore,
\begin{eqnarray}
\frac{d}{dr}\|J(r)\|_{\mathrm{op}}
&\leq&
\ell_r \|J(r)\|_{\mathrm{op}}.
\end{eqnarray}

Applying Grönwall's inequality gives
\begin{eqnarray}
\|J(\tau)\|_{\mathrm{op}}
&\leq&
\exp\!\left(
\int_0^\tau \ell_r\,dr
\right)
=
\kappa.
\end{eqnarray}

Hence,
\begin{eqnarray}
\|\nabla_{\bm{x}(0)} \bm{x}(\tau)\|_{\mathrm{op}}
&\leq&
\kappa.
\end{eqnarray}
\end{proof}

\begin{lemma}\label{lem:onestep}
    Let $L(\theta)$ be defined as in Eq. \eqref{eq: L_theta}. Let $\bm{x}^\star(r)$ denote the oracle trajectory satisfying
    \begin{eqnarray}
        \frac{d}{dr}\bm{x}^\star (r) &=& \bm{u}(\bm{x}^\star(r), r), \qquad \bm{x}^\star(0) = \bm{x}.
    \end{eqnarray}
    For any $\bm{x} \sim \pi_0(\bm{x})$, we have the one-step prediction $\hat{\bm{x}}(\tau) = \bm{x} + \tau \, \bar{\bm{u}}_\theta(\bm{x}, 0, \tau)$ and the true target sample $\bm{x'}(\tau) = \bm{x} + \int_0^\tau \bm{u}(\bm{x}^\star(r), r)dr$, then
    \begin{eqnarray}
        \mathbb{E}_{\bm{x}\sim \pi_0(\bm{x})}\|\hat{\bm{x}}(\tau)- \bm{x'}(\tau)\|^2 &\leq& \kappa^2 m \tau^2 L(\theta)
    \end{eqnarray}
\end{lemma}

\begin{proof}
    Define the trajectory error
    \begin{eqnarray}
        \bm{e}(r) &\triangleq& \hat{\bm{x}}(r) - \bm{x}^\star(r).
    \end{eqnarray}
    Note that the predicted trajectory satisfies
    \begin{eqnarray}
        \frac{d}{dr}\hat{\bm{x}}(r) &=& \hat{u}_{\theta}(\hat{x}(r), r), \qquad \hat{\bm{x}}(0) = \bm{x}.
    \end{eqnarray}

    Substracting the two dynamics gives
    \begin{eqnarray}
        \frac{d}{dr}\bm{e}(r) &=& \hat{u}_{\theta}(\hat{\bm{x}}(r), r) - \bm{u}(\bm{x}^\star(r), r).
    \end{eqnarray}
    Applying Alekseev-Gröbner identity \citep{grobner1960lie}, we obtain
    \begin{eqnarray}
        \bm{e}(\tau) &=& \int_0^\tau \nabla_{\bm{x}(0)} \bm{x}(\tau)\Big(\hat{u}_{\theta}(\hat{\bm{x}}(r), r) - \bm{u}(\bm{x}^\star(r), r)\Big)\,dr,
    \end{eqnarray}
    Applying Lemma \ref{lem:gronwall}, we have
    \begin{eqnarray}
        \|\nabla_{\bm{x}(0)} \bm{x}(\tau)\|_{\mathrm{op}} &\leq& \kappa.
    \end{eqnarray}
    Therefore, 
    \begin{eqnarray}
        \|\hat{\bm{x}}(\tau) - \bm{x'}(\tau)\| &=& \|\bm{e}(\tau)\| \ \leq \ \kappa \int_0^\tau \|\hat{u}_{\theta}(\hat{\bm{x}}(r), r) - \bm{u}(\bm{x}^\star(r), r)\|\,dr
    \end{eqnarray}
    Applying Cauchy-Schwarz inequality gives
    \begin{eqnarray}
        \|\bm{e}(\tau)\|^2 &\leq& \kappa^2 \tau\int_0^\tau \|\hat{u}_{\theta}(\hat{\bm{x}}(r), r) - \bm{u}(\bm{x}^\star(r), r)\|^2\,dr
    \end{eqnarray}

    Taking expectation with respect to $\bm{x}\sim\pi_0(\bm{x})$, we obtain
    \begin{eqnarray}\label{eq:lem4_first_bound}
        \mathbb{E}_{\bm{x}\sim\pi_0(\bm{x})} \|\hat{\bm{x}}(\tau) - \bm{x'}(\tau)\|^2 &\leq& \kappa^2 \tau  \int_0^\tau \mathbb{E}_{\bm{x}\sim\pi_0(\bm{x})} \Big\|\bm{u}_\theta (\bm{x}(r), r) - \frac{d}{dt}\bm{x}^\star(r)\Big\|_2^2 dr \\
        &=& \kappa^2 \tau^2 \, \sum_{i=1}^m \mathbb{E}_{t\sim\mathcal{U}(0,\tau)} \mathbb{E}_{(\bm{x}, \bm{x'})}  \Big\|\bm{u}^i_\theta (\bm{x}(t), t) - \frac{d}{dt}\bm{x}^\star_i(t)\Big\|^2_2\nonumber
    \end{eqnarray}

    We now need to bound the above sum in terms of $L(\theta)$. For any fixed $t\in[0, \tau]$, the pair $(s, t)$ with $s\in [0,t]$ is a subset of all pairs sampled in $L(\theta)$. Therefore, for a process $i$ and fixed $t$:
    \begin{eqnarray*}
        \mathbb{E}_{s\sim\mathcal{U}(0,t)} \mathbb{E}_{(\bm{x}, \bm{x'})} \Big\|\bm{u}^i_\theta (\bm{x}(t), t) - \frac{d}{dt}\bm{x}^\star_i(t)\Big\|^2_2 &\leq& \mathbb{E}_{\substack{s\sim\mathcal{U}(0,\tau)\\t'\sim\mathcal{U}(s,\tau)}}\mathbb{E}_{(\bm{x}, \bm{x'})}\Big\|\bm{u}^i_\theta (\bm{x}(t'), t') - \frac{d}{dt}\bm{x}^\star_i(t')\Big\|^2_2
    \end{eqnarray*}
    Integrating over $t$ from $0$ to $\tau$ and summing over processes $i=1,\ldots, m$ yields
    \begin{eqnarray}
        \int_0^\tau \mathbb{E}_{\bm{x}\sim\pi_0(\bm{x})} \Big\|\bm{u}_\theta (\bm{x}(t), t) - \frac{d}{dt}\bm{x}^\star(t)\Big\|_2^2 dt &\leq& m\tau\, L(\theta).
    \end{eqnarray}
    Substituting this into \eqref{eq:lem4_first_bound} implies
    \begin{eqnarray}
        \mathbb{E}_{\bm{x}\sim\pi_0(\bm{x})} \|\hat{\bm{x}}(\tau) - \bm{x'}(\tau)\|^2 &\leq& \kappa^2 m \tau^2L(\theta)
    \end{eqnarray}
\end{proof}

\subsection{Proof of Theorem~\ref{thm:2}}
We now ready to prove Theorem~\ref{thm:2}

\begin{proof}[Proof of Theorem~\ref{thm:2}]
    By properties of the Wasserstein-2 distance on product measures \citep{villani2009optimal}, we have
    \begin{eqnarray}\label{eq:sum-w2}
    \sum_{i=1}^m W_2^2(\pi_\tau^i, \bar{\pi}_\tau^i)
    \;&\leq&
    W_2^2(\pi_\tau, \bar{\pi}_\tau),
    \end{eqnarray}
    where $\pi_\tau$ and $\bar{\pi}_\tau$ denote the joint distributions over the full multi-physics state.
    
    Since $W_2^2$ is the infimum over all couplings of the expected squared Euclidean cost, for the particular coupling induced by the one-step predictor we obtain
    \begin{eqnarray}\label{eq:joint-w2}
        W_2^2(\pi_\tau, \bar{\pi}_\tau)
        \;&\leq&\;
        \mathbb{E}_{\bm{x} \sim \pi_0(\bm{x})} \bigl[ \|\hat{\bm{x}}(\tau) - \bm{x}'(\tau)\|_2^2 \bigr].
    \end{eqnarray}
    
    From Lemma~\ref{lem:onestep}, we have
    \begin{eqnarray}\label{eq:one-step-bound-final}
        \mathbb{E}_{\bm{x} \sim \pi_0} \bigl[ \|\hat{\bm{x}}(\tau) - \bm{x}'(\tau)\|_2^2 \bigr]
        \;&\leq&
        \kappa^2 m \tau^2 L(\theta).
    \end{eqnarray}

    Combining \eqref{eq:sum-w2}, \eqref{eq:joint-w2}, and \eqref{eq:one-step-bound-final} yields
    \begin{eqnarray}
        \frac{1}{m} \sum_{i=1}^m W_2^2(\pi_\tau^i, \bar{\pi}_\tau^i)
        &\leq&
    \kappa^2 \tau^2 L(\theta).
    \end{eqnarray}

   Using Jensen's inequality gives
    \begin{eqnarray}
        \textsc{err}(\theta) &\leq& \sqrt{\frac{1}{m}\sum_{i=1}^mW_2^2(\pi_\tau^i, \bar{\pi}_\tau^i)} \ \leq \ \kappa \tau \sqrt{L(\theta)}.
    \end{eqnarray}
   
    This concludes the proof.
\end{proof}

\begin{table}[t]
\centering

\resizebox{0.65\textwidth}{!}{
\begin{tabular}{lrrrrr}
\toprule
\textbf{Model} & Grey-Scot & Multiphase & Lotka-V. & Belousov-Z & THM \\
\midrule
Transolver   & 0.42 & 0.42 & 0.42 & 0.42 & 0.42 \\
FNO          & 4.47  & 4.47  & 0.58 & 0.58 & 8.41  \\
UFNO         & 6.73 & 6.73 & 1.31  & 1.31  & 10.04 \\
CMWNO        & 2.0  & 2.0  & 1.0  & 1.5  & 5.0  \\
COMPOL-RNN       & 2.42 & 2.42 & 2.50 & 3.07 & 10.62 \\
COMPOL-ATN       & 2.40 & 2.40 & 1.49 & 2.10 & 10.57 \\
DiffusionPDE & 15.5 & 15.5 & 10.9 & 10.9 & 15.5 \\
CViT-B       & 23.5 & 23.5 & 21.5  & 21.5  & 23.6 \\
\midrule
\textbf{Ours} & 10.4 & 10.4 & 8.9 & 13.4 & 25.5 \\
\bottomrule
\end{tabular}
}
\vspace{2mm}
\caption{Number of model parameters (M) of baselines across different benchmark datasets.}
\label{tab:param_count}
\end{table}

\begin{table}[t]
\centering

\resizebox{0.65\textwidth}{!}{
\begin{tabular}{lrrrrr}
\toprule
\textbf{Model} & Grey-Scot & Multiphase  & Belousov-Z & Lotka-V. & THM \\
\midrule
FNO          & 0.278  & 0.308  & 0.132 & 0.142 & 0.271  \\
UFNO         & 0.836 & 0.839 &  0.492  & 0.497  & 0.835 \\
COMPOL-RNN       & 0.910 & 0.880 & 0.608 & 0.502 & 1.818 \\
COMPOL-ATN       & 1.226 & 1.190 & 0.661 & 0.468 & 2.725 \\
CViT-B       & 0.539 & 0.214 & 0.149  & 0.208  & 0.210 \\
CMWNO        & 11.471  & 9.449  & 7.912  & 11.284  & 19.870  \\
DiffusionPDE & 58.013 & 93.442 & 78.458 & 66.134 & 60.065 \\
\midrule
\textbf{Ours} & 3.117 & 2.926 & 5.328 & 3.601 & 4.943 \\
\bottomrule
\end{tabular}
}
\vspace{2mm}
\caption{The wall clock time (ms) of baselines across different benchmark datasets.}
\label{tab:infer_time}
\end{table}

\begin{figure}
    \centering
    \includegraphics[width=1.0\linewidth]{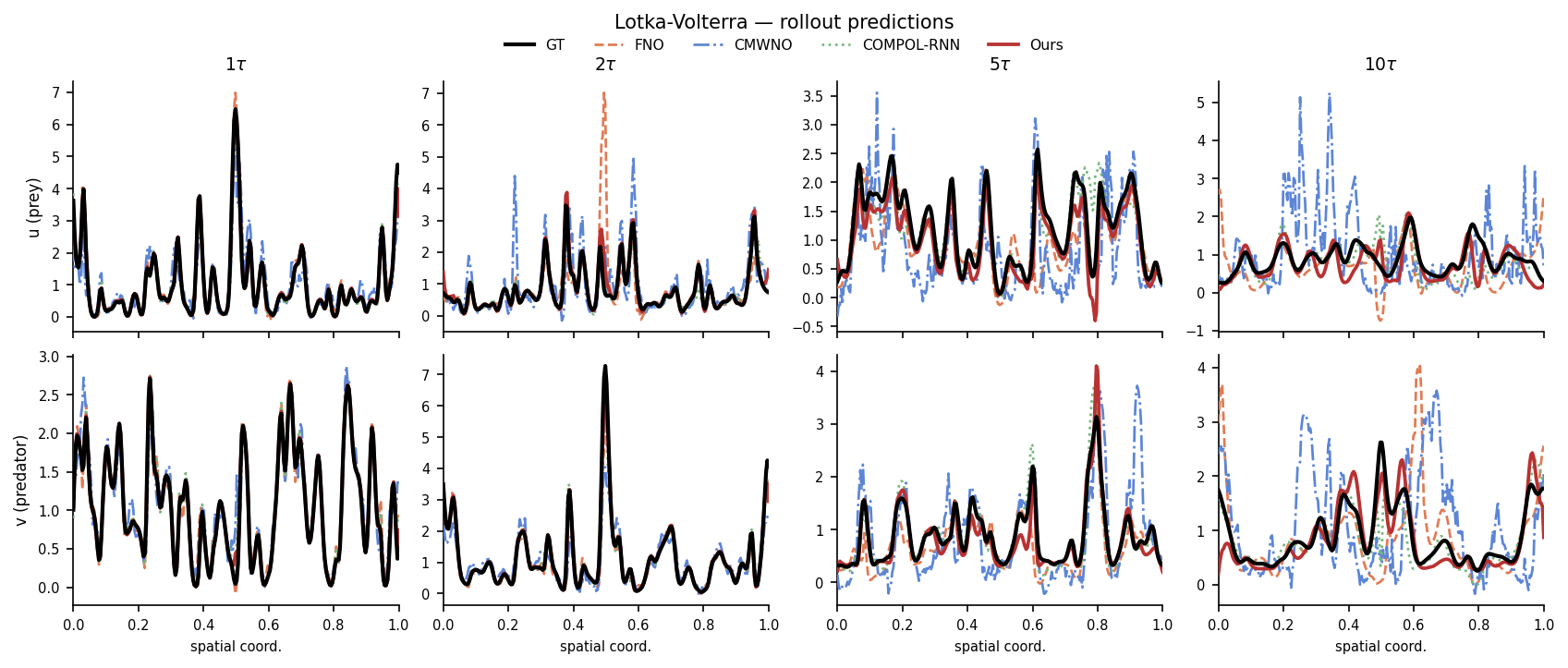}
    \caption{The prediction of \textsc{coupled-flow}-GP (Ours) and other baselines vs. Ground Truth (GT) on Lotka--Volterra across different rollout prediction steps.}
    \label{fig:lv_pred_rollout}
\end{figure}

\section{Detailed Experiment Results and Ablation Studies}
\label{appendix:Experiment_detailed_results}

In this section, we provide the detailed experiment results and additional ablation studies that support the empirical analysis in Section~\ref{sec:experiments}. 
Unless otherwise stated, all reported numbers are relative $L_2$ prediction errors averaged over five independent runs, and the uncertainty term denotes the standard deviation across runs. 
The main results in Section~\ref{sec:experiments} are reported in the scaled form $(\times 10^{-2})$, while the tables in this appendix report the corresponding raw relative $L_2$ values.

\textbf{Detailed single-step prediction results.}
Table~\ref{tab:nrmse_main_std} reports the complete single-step prediction results on the five coupled multi-physics benchmarks, including Lotka--Volterra, Belousov--Zhabotinsky, Gray--Scott, Multiphase, and Thermal-Hydro-Mechanical systems. 
The results provide the mean and standard deviation corresponding to Table~\ref{tab:main_results}. 
Across both training-data regimes, \textsc{coupled-flow}-GP consistently achieves the best or second-best performance. 
In particular, \textsc{coupled-flow}-GP obtains the strongest performance on most settings, while \textsc{coupled-flow}-Linear also remains highly competitive. 
This confirms that the improvement does not only come from a larger neural architecture, but from the coupled transport formulation itself.

\begin{table}[t]
\centering
\resizebox{\textwidth}{!}{%
\begin{tabular}{lccccc}
\toprule
\textbf{Hyperparameter} & Grey-Scott & Lotka-V. & Multiphase & Belousov-Z & THM \\
\midrule
Architecture & Unet & Unet & Unet & Unet & Unet \\
\# nets & 2 & 2 & 2 & 3 & 5 \\
Params/net & $\sim$5.1M & $\sim$4.5M & $\sim$5.1M & $\sim$4.5M & $\sim$5.1M \\
Total params & $\sim$10.2M & $\sim$9M & $\sim$10.2M & $\sim$13.5M & $\sim$25.5M \\
Batch size & 32 & 32 & 32 & 32 & 32 \\
Learning rate & $6\times10^{-4}$ & $6\times10^{-4}$ & $6\times10^{-4}$ & $6\times10^{-4}$ & $6\times10^{-4}$ \\
Epochs & 500 & 500 & 500 & 500 & 500 \\
Warmup epochs & 100 & 100 & 100 & 100 & 100 \\
LR schedule & linear warmup &  linear warmup  & linear warmup & linear warmup & linear warmup \\
Optimizer & Adam &  Adam  & Adam & Adam & Adam \\
Dropout & 0.2 & 0.2 & 0.2 & 0.2 & 0.2 \\
EMA decay & 0.9999 & 0.9999  & 0.9999  & 0.9999  & 0.9999  \\
GP kernel & SE & SE & SE & SE & SE \\
GP length scale $\ell_0$&  1 & 1 & 1 & 1 & 1 \\
Train ratio & 0.8 & 0.8 & 0.8 & 0.8 & 0.8 \\
Val ratio & 0.2 & 0.2 & 0.2 & 0.2 & 0.2 \\
\bottomrule
\end{tabular}%
}
\vspace{2mm}
\caption{Hyperparameter settings across benchmark datasets. The Adam~\citep{kingma2017adammethodstochasticoptimization} optimizer is referred to as Adam with $(\beta_1=0.9,\beta_2=0.999)$. SE is referred to as the Squared-Exponential kernel.}
\label{tab:hyperparameters}
\end{table}

\textbf{Results on standard PDE benchmarks.}
Table~\ref{tab:rd_ns_comparison_std} provides the complete results with standard deviations on Reaction-Diffusion-2D and Navier-Stokes-2D. 
These benchmarks complement the coupled PDE datasets used in prior coupled-operator work and evaluate whether the proposed formulation remains effective on standardized operator-learning benchmarks. 
On Reaction-Diffusion-2D, \textsc{coupled-flow}-GP achieves performance close to the strongest CViT-B baseline while substantially improving over FNO, UFNO, CMWNO, COMPOL, DiffusionPDE, and MeanFlow. 
On Navier-Stokes-2D, \textsc{coupled-flow}-GP achieves the best performance among all compared methods. 
\textsc{coupled-flow}-Linear is also competitive with CViT-B, suggesting that the one-step coupled average-velocity parameterization is already effective even without the GP path, while the GP path provides additional gains.

\textbf{Rollout prediction.}
Tables~\ref{tab:rollout_lv_std} and ~\ref{tab:rollout_gs_std} report recursive rollout prediction on Lotka--Volterra and Gray--Scott. 
At each rollout step, the model prediction is recursively fed back as the next input, making this evaluation more challenging than single-step prediction. 
On Lotka--Volterra, \textsc{coupled-flow}-GP achieves the best performance at $1\tau$, $2\tau$, and $10\tau$, while \textsc{coupled-flow}-Linear obtains the best performance at $5\tau$. 
On Gray--Scott, \textsc{coupled-flow}-GP consistently achieves the lowest error across all rollout horizons. 
These results indicate that the learned one-step transport is stable under recursive application and does not merely optimize short-horizon prediction.

\textbf{Length-scale Sensitivity.}
Table~\ref{tab:lengthscale_sens_std} studies the sensitivity of \textsc{coupled-flow}-GP to the squared-exponential kernel length-scale. 
We vary the length-scale around the default value $\ell_0=1$ and evaluate the resulting performance on Lotka--Volterra and Gray--Scott. 
The performance remains stable across a broad range of length scales. 
For Lotka--Volterra, the error changes only slightly across the tested range, while for Gray--Scott the default setting achieves the best result. 
This shows that \textsc{coupled-flow}-GP does not require delicate length-scale tuning to obtain competitive performance.

\textbf{Kernel Sensitivity.}~Table~\ref{tab:kernel_sensitivity} compares different GP kernels, including squared-exponential, Mat\'ern-$3/2$, Mat\'ern-$5/2$, and rational quadratic kernels. 
All GP kernels perform competitively with or better than the linear interpolant. 
Different kernels achieve the best results on different datasets: Mat\'ern-$3/2$ performs best on Lotka--Volterra, while the rational quadratic kernel performs best on Gray--Scott. 
This indicates that the empirical gains are not tied to one specific covariance function. 
Instead, the improvement mainly comes from replacing independent linear interpolation with a correlated GP-based path construction.

\textbf{Coefficient-shift generalization.}
Table~\ref{tab:ood_coeff} evaluates out-of-distribution generalization under shifted coefficient conditions on Lotka--Volterra and Gray--Scott. 
\textsc{coupled-flow}-GP achieves the best OOD performance on both datasets, while \textsc{coupled-flow}-Linear remains the second-best method in most settings. 
On Lotka--Volterra, \textsc{coupled-flow}-GP improves over CViT-B under the OOD coefficient setting. 
On Gray--Scott, both \textsc{coupled-flow} variants substantially outperform the other baselines under coefficient shift. 
These results suggest that the proposed coupled transport objective improves robustness when the test dynamics deviate from the training coefficient distribution.

\begin{table}[t]
\centering
\small
\begin{tabular}{lcc}
\toprule
Method & MPF-ID & MPF-OOD \\
\midrule
DiffusionPDE & 0.0241 & 0.0318 \\
\textsc{CoupledFlow}-Linear & 0.0090 & 0.0240 \\
\textsc{CoupledFlow}-GP &
\textbf{0.0061} & \textbf{0.0190} \\
\bottomrule
\end{tabular}
\vspace{3mm}
\caption{Comparison of relative $L_2$ errors
under in-distribution (ID) and out-of-distribution (OOD)
settings on Multiphase Flow (MPF), using approximately
matched model capacity (10M parameters).}
\label{tab:mpf_ood}
\end{table}

\begin{figure}
    \centering
    \includegraphics[width=0.9\linewidth]{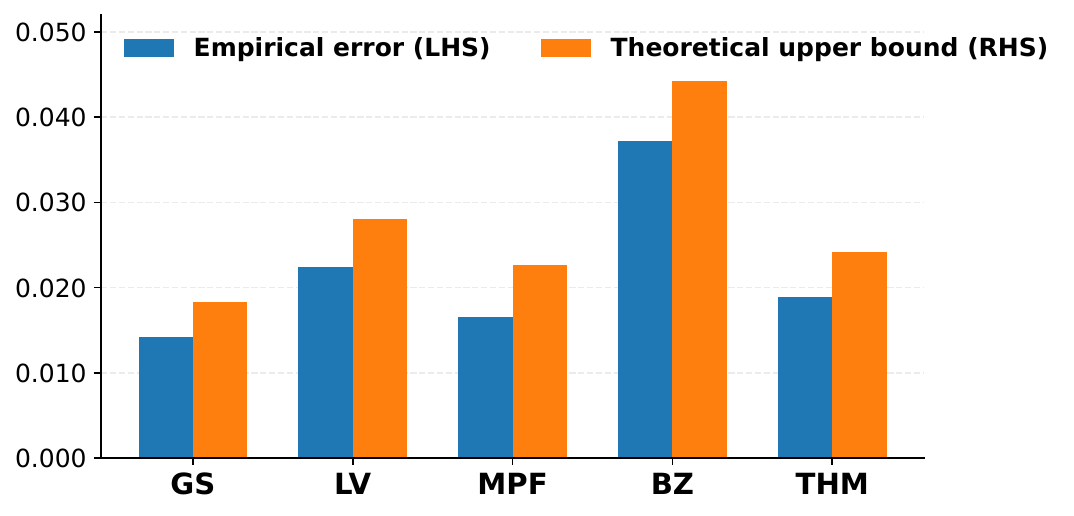}
    \caption{Empirical evaluation of the generalization bound in
Theorem~2 across five coupled multi-physics benchmarks.
We compare the estimated prediction error
$\mathrm{ERR}(\theta)$ (LHS) with its theoretical upper
bound $\kappa\tau\sqrt{\mathcal{L}(\theta)}$ (RHS).
The empirical results satisfy the inequality across
all datasets, with relatively small differences between
the two quantities, suggesting that the bound is informative.}
    \label{fig:theorem2_bound}
\end{figure}

\textbf{Empirical Verification of the Generalization Bound. } Theorem~\ref{thm:2} establishes an upper bound on the average
Wasserstein-2 prediction error of \textsc{CoupledFlow}:
\begin{eqnarray}
\textsc{err}(\theta) &\leq& \kappa \tau \sqrt{L(\theta)} , \nonumber
\end{eqnarray}
where $\kappa = \exp(\int_0^\tau \ell_r\,dr)$ and
$\ell_r$ denotes the Lipschitz constant of the induced
instantaneous velocity. Importantly, the objective function $L(\theta)$ in Eq.~\eqref{eq: L_theta}
is not restricted to linear interpolants.
In the GP variant, we use the coupled transport paths
constructed in Algorithm~\ref{alg:training}: one randomly selected process
is initialized using linear interpolation, while the
remaining processes are constructed using the GP
conditional means. To assess the informativeness of this bound, we empirically
estimate its left-hand side (LHS) and right-hand side (RHS)
on five multi-physics benchmarks. The path-matching loss $L(\theta)$
 is estimated with respect to the learned theta using Eq.~\eqref{eq: L_theta} with 1000 sampled time points on a test set with 256 data points. The computation is also averaged over 5 independent runs. For each sampled time 
$r$, we estimate the local Lipschitz constant as $\ell_{r} = \max_{\bm{x}} \|\nabla_{\bm{x} \in \mathcal{S}_{r}} \bm{u}_{\theta}(\bm{x}, r) \|_2$, where 
 $\mathcal{S}_{r}$ contains the states sampled from the held-out trajectories at time $r$. We then estimate the Jacobian spectral norm $|\cdot|_2$ following standard practice~\citep{scaman2019lipschitzregularitydeepneural} for estimating the network's local Lipschitz constant. Fig.~\ref{fig:theorem2_bound}
visualizes their comparison, which shows that across all five datasets, the empirical LHS remains below
the estimated RHS. The relative gaps, defined as
$(\mathrm{RHS}-\mathrm{LHS})/\mathrm{LHS}$, range from
$18.8\%$ to $37.0\%$, with an average of $27.6\%$.
These results suggest that the bound remains informative
for the GP-based formulation across tested coupled
PDE systems.

\textbf{Sparse Gaussian Process Baseline.}~To further investigate whether the performance gains of
\textsc{CoupledFlow}-GP arise simply from the use of Gaussian
Processes, we compare our method against a sparse Gaussian
Process (SGP)~\citep{NghiaICML15,NghiaICML16,NghiaAAAI17,NghiaAAAI19,NghiaNIPS20} baseline across five coupled multi-physics
benchmarks. Specifically, the SGP directly learns the mapping from the
initial coupled state $\bm{x}$, obtained by concatenating
all physical processes, to the corresponding future state
$\bm{x}'$ at the target prediction horizon.
We adopt a sparse GP instead of a full GP due to the large
number of training pairs in the Belousov--Zhabotinsky,
Multiphase Flow, and THM datasets, which makes exact GP inference
computationally expensive.
For these three datasets, we use $M=1500$ inducing points,
yielding a computational complexity of
$\mathcal{O}(M^3 + M^2 N)$, where $N$ denotes the number
of training samples. As shown in Table~\ref{tab:sgp_baseline},
\textsc{CoupledFlow}-GP consistently outperforms
the SGP baseline across all five benchmarks.
Moreover, even \textsc{CoupledFlow}-Linear, which
does not employ GP-based interpolants, substantially
outperforms SGP.
These results suggest that the superior performance
of \textsc{CoupledFlow}-GP cannot be attributed
solely to the use of Gaussian Processes.
Instead, the proposed coupled mean-flow formulation,
together with GP-based correlated intermediate paths,
provides a more effective mechanism for learning
cross-physics dependencies than direct GP regression.

\begin{table}[t]
\centering
\small
\begin{tabular}{llccc}
\toprule
System & Method & Overall $L_2$ & BC Error & IC Error \\
\midrule
\multirow{5}{*}{MPF}
& FNO
    & 0.02687 & 0.04363 & 0.02766 \\
& CViT
    & 0.14251 & 0.16921 & 0.11824 \\
& Transolver
    & 0.01325 & 0.02340 & 0.01618 \\
& COMPOL
    & 0.09545 & 0.12070 & 0.09705 \\
& \textsc{CoupledFlow}
    & \textbf{0.00575} & \textbf{0.00741} & \textbf{0.00539} \\
\midrule
\multirow{5}{*}{THM}
& FNO
    & 0.00198 & 0.00216 & 0.00223 \\
& CViT
    & 0.02945 & 0.03293 & 0.03389 \\
& Transolver
    & 0.00184 & \textbf{0.00183} & \textbf{0.00071} \\
& COMPOL
    & 0.00171 & 0.00187 & 0.00170 \\
& \textsc{CoupledFlow}
    & \textbf{0.00108} & \textbf{0.00130} & 0.00123 \\
\bottomrule
\end{tabular}
\hspace{2mm}
\caption{Comparison of overall relative $L_2$ prediction error,
boundary condition (BC) error, and initial condition (IC) error
on Multiphase Flow (MPF) and Thermal-Hydro-Mechanical (THM)
benchmarks using 512 training samples.
The best results are highlighted in bold.}
\label{tab:additional_metrics}
\end{table}

\paragraph{Boundary and Initial-Condition Prediction Errors.}
To further assess the predictive performance of \textsc{CoupledFlow},
we evaluate additional metrics, including boundary condition (BC)
and initial condition (IC) errors, alongside the overall relative
$L_2$ prediction error. We conduct this evaluation on the Multiphase
Flow (MPF) and Thermal-Hydro-Mechanical (THM) benchmarks using
512 training samples, with fresh training runs for each method. As shown in Table~\ref{tab:additional_metrics},
\textsc{CoupledFlow} consistently achieves the lowest overall
prediction error and boundary condition error on both benchmarks.
On MPF, our method also obtains the lowest IC error,
substantially outperforming all competing baselines across
the three metrics. On THM, \textsc{CoupledFlow} achieves
the lowest overall and boundary errors, while Transolver
obtains the lowest IC error. These results demonstrate that the predictive improvements
of \textsc{CoupledFlow} extend beyond the overall solution
field to boundary and initial-condition regions.
In particular, the consistently lower boundary errors
suggest that our method more accurately captures
solution behavior near domain boundaries.

\begin{table}[t]
\centering
\small
\resizebox{\linewidth}{!}{
\begin{tabular}{lccccc}
\toprule
Method & Lotka-V. & Belousov-Z. & Gray-Scott
       & Multiphase & THM \\
\midrule
\textsc{CoupledFlow}-Linear
    & 0.0234 & 0.0015 & 0.0023 & 0.0090 & 0.0011 \\

\textsc{CoupledFlow}-GP
    & \textbf{0.0216}
    & \textbf{0.0011}
    & \textbf{0.0019}
    & \textbf{0.0061}
    & \textbf{0.0006} \\

Sparse GP
    & 0.3554 & 0.0222 & 0.1162 & 0.0781 & 0.1443 \\
\bottomrule
\end{tabular}
}
\caption{Comparison of relative $L_2$ prediction errors
between \textsc{CoupledFlow} and the sparse Gaussian
Process (SGP) baseline across five coupled multi-physics
benchmarks. The best results are highlighted in bold.}
\label{tab:sgp_baseline}
\end{table}

\begin{table}[t]
\centering
\resizebox{0.65\textwidth}{!}{
\begin{tabular}{lcc}
\toprule
Method & Lotka-V. & Gray-Scott \\
\midrule
\textsc{coupled-flow}-Linear & 0.0234 $\pm$ 1e-4 & 0.0023 $\pm$ 1e-5 \\
\midrule
\textsc{coupled-flow}-GP ($0.25 \ell_0$) & 0.0226 $\pm$ 2e-4 & 0.0022 $\pm$ 3e-5 \\
\textsc{coupled-flow}-GP ($0.5 \ell_0$)  & 0.0218 $\pm$ 8e-5 & 0.0023 $\pm$ 2e-5\\
\textsc{coupled-flow}-GP ($1.0 \ell_0$)  & \textcolor{blue}{\textbf{0.0216 $\pm$ 2e-5}} & \textbf{0.0019 $\pm$ 3e-5} \\
\textsc{coupled-flow}-GP ($2.0 \ell_0$)  & \textbf{0.0215 $\pm$ 1e-4} & 0.0021 $\pm$ 4e-5 \\
\textsc{coupled-flow}-GP ($4.0 \ell_0$)  & 0.0217 $\pm$ 5e-5 & \textcolor{blue}{\textbf{0.0020 $\pm$ 5e-5}}\\
\bottomrule
\end{tabular}
}
\vspace{2mm}
\caption{
Sensitivity of \textsc{coupled-flow}-GP to the squared-exponential kernel length-scale.
We report relative $L_2$ errors ($\times 10^{-2}$) using $512$ training samples with $\ell_0=1$ as the practical implementation.
}
\label{tab:lengthscale_sens_std}
\end{table}

\begin{table}[t]
\centering
\resizebox{0.65\textwidth}{!}{
\begin{tabular}{lcc}
\toprule
Kernel & Lotka-V. & Gray-Scott \\
\midrule
Our-Linear & 0.0234 $\pm$ 1e-4 & 0.0023 $\pm$ 1e-5 \\
\midrule
Squared-Exponential & \textcolor{blue}{\textbf{0.0216 $\pm$ 2e-5}} & \textcolor{blue}{\textbf{0.0019 $\pm$ 3e-5}} \\
Matérn-$3/2$ & \textbf{0.0214 $\pm$ 3e-5} & 0.0022 $\pm$ 2e-5 \\
Matérn-$5/2$ & 0.0230 $\pm$ 1e-4 & 0.0023 $\pm$ 7e-5 \\
Rational Quadratic & 0.0223 $\pm$ 2e-5 & \textbf{0.0018 $\pm$ 1e-4} \\
\bottomrule
\end{tabular}
}
\vspace{2mm}
\caption{
Kernel sensitivity of \textsc{coupled-flow}-GP. We report relative $L_2$ errors (mean $\pm$ std) using $512$ training samples. All GP kernels use the same
default length-scale $\ell_0 =1$.
}
\label{tab:kernel_sensitivity}
\end{table}

\begin{table}[!htbp]
\centering
\renewcommand{\arraystretch}{1.2}
\setlength{\tabcolsep}{0.35em}
\resizebox{\columnwidth}{!}{%
\begin{tabular}{@{}lcccccc@{}}
\hline
\textbf{Method} & \textit{Train Samples} & \textit{Lotka-V.} & \textit{Belousov-Z.} & \textit{Grey-Scott} & \textit{Multiphase} & \textit{THM} \\
\hline
\multirow{2}{*}{$\text{FNO}_{c}$} & 512 & 0.1229 $\pm$4e-5 & 0.0023 $\pm$3e-5 & 0.0088 $\pm$4e-5 & 0.0229 $\pm$2e-4 & 0.0015 $\pm$2e-5 \\
& 1024 & 0.0955 $\pm$6e-5 & 0.0016 $\pm$3e-5 & 0.0055 $\pm$3e-6 & 0.0291 $\pm$1e-5 & 0.0011 $\pm$1e-6 \\
\hline
\multirow{2}{*}{CMWNO} & 512 & 0.0735 $\pm$3e-5 & 0.0021 $\pm$2e-5 & 0.0083 $\pm$5e-4 & 0.0256 $\pm$2e-5 & 0.0013 $\pm$4e-5 \\
& 1024 & 0.0665 $\pm$1e-5 & 0.0014 $\pm$6e-5 & 0.0043 $\pm$3e-5 & 0.0193 $\pm$6e-5 & 0.0009 $\pm$5e-5 \\
\hline
\multirow{2}{*}{$\text{UFNO}_{c}$} & 512 & 0.1163 $\pm$1e-4 & 0.0026 $\pm$6e-6 & 0.0055 $\pm$5e-5 & 0.0448 $\pm$1e-5 & 0.0013 $\pm$3e-5 \\
& 1024 & 0.0873 $\pm$8e-5 & 0.0017 $\pm$2e-5 & 0.0036 $\pm$6e-6 & 0.0387 $\pm$1e-5 & 0.0008 $\pm$4e-5 \\
\hline
\multirow{2}{*}{$\text{Transolver}_{c}$} & 512 & 0.1819 $\pm$1e-5 & 0.0597 $\pm$7e-6 & 0.0054 $\pm$9e-6 & 0.0128 $\pm$4e-5 & 0.0018 $\pm$1e-5 \\
& 1024 & 0.1767 $\pm$1e-5 & 0.0188 $\pm$2e-5 & 0.0049 $\pm$2e-5 & 0.0088 $\pm$3e-5 & 0.0016 $\pm$5e-5 \\
\hline
\multirow{2}{*}{$\text{DeepONet}_{c}$} & 512 & 0.5546 $\pm$7e-5 & 0.0111 $\pm$1e-5 & 0.3130 $\pm$1e-5 & 0.3120 $\pm$1e-5 & 0.3136 $\pm$1e-5 \\
& 1024 & 0.5483 $\pm$6e-5 & 0.0099 $\pm$1e-5 & 0.3099 $\pm$1e-5 & 0.2819 $\pm$7e-5 & 0.3150 $\pm$1e-5 \\
\hline
\multirow{2}{*}{CViT-B} & 512 & \textbf{0.0163$\pm$6e-4} & 0.0104$\pm$7e-4 & 0.0209$\pm$2e-3 & 0.0212$\pm$4e-4 & 0.0127$\pm$6e-4 \\
& 1024 & \textbf{0.0077$\pm$3e-4} & 0.0031$\pm$1e-5 & 0.0074$\pm$5e-4 & 0.0139$\pm$4e-4 & 0.0051$\pm$1e-5 \\
\hline
\multirow{2}{*}{$\text{DiffusionPDE}_{c}$} & 512 & 0.0441 $\pm$ 3e-4 & 0.0033 $\pm$ 2e-5 & 0.0026$\pm$ 1e-4 & 0.0361 $\pm$ 5e-5 & 0.0021 $\pm$ 3e-5 \\
& 1024 & 0.0310 $\pm$ 3e-5 & 0.0053 $\pm$ 7e-5 & 0.0021 $\pm$ 5e-4 & 0.0025 $\pm$ 5e-5 & 0.0016 $\pm$ 2e-5 \\
\hline
\multirow{2}{*}{COMPOL-RNN} & 512 & 0.1043 $\pm$2e-5 & 0.0017 $\pm$2e-5 & 0.0066 $\pm$1e-4 & 0.0221 $\pm$1e-5 & 0.0014 $\pm$3e-5 \\
& 1024 & 0.0802 $\pm$4e-5 & 0.0015 $\pm$8e-6 & 0.0050 $\pm$4e-5 & 0.0644 $\pm$3e-5 & 0.0009 $\pm$4e-5 \\
\hline
\multirow{2}{*}{COMPOL-ATN} & 512 & 0.1132 $\pm$1e-4 & 0.0018 $\pm$4e-5 & 0.0061 $\pm$3e-5 & 0.0217 $\pm$3e-5 & 0.0014 $\pm$3e-5 \\
& 1024 & 0.0858 $\pm$5e-5 & 0.0015 $\pm$3e-5 & 0.0048 $\pm$2e-5 & 0.0676 $\pm$3e-5 & 0.0010 $\pm$2e-5 \\
\hline
\multirow{2}{*}{$\text{MeanFlow}_{c}$} & 512 & 0.1650 $\pm$4e-3 & 0.0018 $\pm$6e-5 & 0.0033 $\pm$5e-5 & 0.0112 $\pm$4e-5 & 0.0017 $\pm$3e-5 \\
& 1024 & 0.1320 $\pm$5e-3 & 0.0017 $\pm$2e-5 & 0.0029 $\pm$6e-5 & 0.0333 $\pm$ 7e-5 & 0.0013 $\pm$4e-5 \\
\hline
\hline
\multirow{2}{*}{\textsc{coupled-flow}-Linear} & 512 & 0.0234 $\pm$1e-4 & \textcolor{blue}{\textbf{0.0015 $\pm$5e-5}} & \textcolor{blue}{\textbf{0.0023 $\pm$1e-5}} & \textcolor{blue}{\textbf{0.0090 $\pm$5e-5}} & \textcolor{blue}{\textbf{0.0011 $\pm$6e-5}} \\
& 1024 & 0.0129 $\pm$4e-5 & \textcolor{blue}{\textbf{0.0009 $\pm$2e-5}} & \textcolor{blue}{\textbf{0.0018 $\pm$5e-5}} & \textbf{0.0009 $\pm$ 4e-5} & \textcolor{blue}{\textbf{0.0009 $\pm$2e-5}} \\
\hline
\multirow{2}{*}{\textsc{coupled-flow}-GP} & 512 & \textcolor{blue}{\textbf{0.0216 $\pm$2e-5}} & \textbf{0.0011 $\pm$4e-5} & \textbf{0.0019 $\pm$3e-5} & \textbf{0.0061 $\pm$1e-5} & \textbf{0.0006 $\pm$2e-5} \\
& 1024 & \textcolor{blue}{\textbf{0.0121 $\pm$3e-5}} & \textbf{0.0006 $\pm$2e-5} & \textbf{0.0015 $\pm$5e-5} & 0.0011 $\pm$1e-5 & \textbf{0.0005 $\pm$2e-5} \\
\hline
\end{tabular}}
\vspace{1em}
\caption{\small Comparison of relative $L_2$ prediction errors (mean $\pm$ std) of our method with different neural operators and generative baselines across different coupled physics datasets. \textbf{Bold} and \textcolor{blue}{\textbf{Blue}} values indicate the best and second-best performance, respectively.}
\label{tab:nrmse_main_std}
\end{table}

\begin{table}[h]
\centering
\begin{tabular}{lcccc}
\toprule
\textbf{Method} & \multicolumn{4}{c}{\textbf{Time-Step}} \\
\cmidrule(lr){2-5}
 & \textbf{1 $\tau$} & \textbf{2 $\tau$} & \textbf{5 $\tau$} & \textbf{10 $\tau$} \\
\midrule
$\text{MeanFlow}_{c}$      & 0.7420 $\pm$ 2e-3  & 1.3310 $\pm$ 5e-3 & 2.3890 $\pm$ 1e-2 & 4.939 $\pm$ 3e-2\\
$\text{FNO}_{c}$          & 0.1240 $\pm$ 3e-4  & 0.2610 $\pm$ 8e-4  & 0.4000 $\pm$ 1e-3  & 0.791 $\pm$ 2e-3 \\
COMPOL-RNN   & 0.1090 $\pm$ 2e-4  & 0.2260 $\pm$ 5e-4  & 0.3780 $\pm$ 1e-3  & 0.533 $\pm$ 3e-3 \\
COMPOL-ATN   & 0.1180 $\pm$ 2e-4 & 0.2890 $\pm$ 7e-4 & 0.3640 $\pm$ 2e-3  & 0.580 $\pm$ 4e-3 \\
CMWNO        & 0.0735 $\pm$ 1e-4 & 0.4528 $\pm$ 8e-4 & 0.7176 $\pm$ 2e-3 & 1.030 $\pm$ 5e-3\\
\midrule
\textsc{coupled-flow}-GP       & \textbf{0.0216 $\pm$ 2e-4} & \textbf{0.1330 $\pm$ 5e-4} & 0.2510 $\pm$ 7e-4 & \textbf{0.3650 $\pm$ 1e-3} \\
\textsc{coupled-flow}-Linear   & 0.0340 $\pm$ 1e-4  & 0.1450 $\pm$ 3e-4  & \textbf{0.2351 $\pm$ 7e-4}  & 0.3772 $\pm$ 2e-3 \\
\bottomrule
\end{tabular}
\vspace{2mm}
\caption{Comparison of relative $L_2$ error rollout prediction (mean $\pm$ std) of our method with different baselines on Lotka-V dataset. \textbf{Bold} and \textcolor{blue}{\textbf{Blue}} values indicate the best and second-best performance, respectively.}
\label{tab:rollout_lv_std}
\end{table}

\begin{table}[h]
\centering
\begin{tabular}{lcccc}
\toprule
\textbf{Method} & \multicolumn{4}{c}{\textbf{Time-Step}} \\
\cmidrule(lr){2-5}
 & \textbf{1$\tau$} & \textbf{2$\tau$} & \textbf{5$\tau$} & \textbf{10$\tau$} \\
\midrule
$\text{MeanFlow}_{c}$      & 0.0033 $\pm$ 1e-4 & 0.0102 $\pm$ 2e-4 & 0.0262 $\pm$ 7e-4 & 0.0591 $\pm$ 1e-3 \\
$\text{FNO}_{c}$          & 0.0088 $\pm$ 2e-4 & 0.0139 $\pm$ 7e-4 & 0.0184 $\pm$ 1e-3 & 0.0368 $\pm$ 3e-3 \\
COMPOL-RNN   & 0.0066 $\pm$ 1e-4 & 0.0134 $\pm$ 4e-4 & 0.0121 $\pm$ 1e-3 & 0.0287 $\pm$ 2e-3 \\
COMPOL-ATN   & 0.0051 $\pm$ 1e-4 & 0.0133 $\pm$ 5e-4 & 0.0119 $\pm$ 1e-3 & 0.0211 $\pm$ 2e-3\\
CMWNO      & 0.0083 $\pm$ 2e-4 & 0.0143 $\pm$ 6e-4 & 0.0196 $\pm$ 2e-3 & 0.0226 $\pm$ 4e-3 \\
\midrule
\textsc{coupled-flow}-GP       & \textbf{0.0019 $\pm$ 1e-4} & \textbf{0.0025 $\pm$ 3e-4} & \textbf{0.0069 $\pm$ 8e-4} & \textbf{0.0167 $\pm$ 1e-3} \\
\textsc{coupled-flow}-Linear   & 0.0023 $\pm$ 1e-4 & 0.0029 $\pm$ 2e-4 & 0.0071 $\pm$ 1e-3 & 0.0168 $\pm$ 2e-3 \\
\bottomrule
\end{tabular}
\vspace{2mm}
\caption{Comparison of relative $L_2$ error rollout prediction (mean $\pm$ std) of our method with different baselines on GrayScott dataset. \textbf{Bold} and \textcolor{blue}{\textbf{Blue}} values indicate the best and second-best performance, respectively.}
\label{tab:rollout_gs_std}
\end{table}

\begin{table}[t]
\centering
\small
\begin{tabular}{lcc}
\toprule
\textbf{Method} 
& \textbf{Reaction-Diffusion-2D} 
& \textbf{Navier-Stokes-2D} \\
\midrule
$\text{MeanFlow}_{c}$      & 0.0951 $\pm$ 4e-4 & 0.1613 $\pm$ 1e-3 \\
$\text{FNO}_{c}$          & 0.1234 $\pm$ 1e-3 & 0.0912 $\pm$ 7e-4 \\
$\text{UFNO}_{c}$          & 0.1233 $\pm$ 7e-4 & 0.0947 $\pm$ 3e-4 \\
COMPOL-RNN   & 0.0828 $\pm$ 3e-4 & 0.0825 $\pm$ 6e-4 \\
COMPOL-ATN   & 0.0827 $\pm$ 4e-4 & 0.0821 $\pm$ 1e-3 \\
CMWNO        & 0.0845 $\pm$ 7e-4 & 0.0842 $\pm$ 2e-4 \\
DiffusionPDE        & 0.1265 $\pm$1e-3 & 0.0884 $\pm$ 6e-4 \\
CViT-B & \textbf{0.0158 $\pm$ 5e-4} & \textcolor{blue}{\textbf{0.0395 $\pm$ 2e-4}} \\
\midrule
\textsc{coupled-flow}-GP       & \textcolor{blue}{\textbf{0.0163 $\pm$ 3e-4}} & \textbf{0.0345 $\pm$ 4e-4} \\
\textsc{coupled-flow}-Linear   & 0.0174 $\pm$ 1e-4 & 0.0398 $\pm$ 5e-4\\
\bottomrule
\end{tabular}
\vspace{2mm}
\caption{Comparison of relative $L_2$ prediction error (mean $\pm$ std) on Reaction-Diffusion-2D and Navier-Stokes-2D. \textbf{Bold} values indicate the best performance for each case. \textcolor{blue}{\textbf{Blue}} values indicate the second-best performance for each case.}
\label{tab:rd_ns_comparison_std}
\end{table}

\begin{table}[t]
\centering
\resizebox{0.75\textwidth}{!}{
\begin{tabular}{lcccc}
\toprule
\multirow{2}{*}{Method} 
& \multicolumn{2}{c}{Lotka--Volterra} 
& \multicolumn{2}{c}{Gray--Scott} \\
\cmidrule(lr){2-3} \cmidrule(lr){4-5}
& ID coeff. & OOD coeff. 
& ID coeff. & OOD coeff. \\
\midrule
FNO$_c$        & 0.1229 $\pm$4e-5 & 0.1815 $\pm$ 1e-4 & 0.0088 $\pm$ 4e-5 & 0.1015 $\pm$ 1e-4 \\
CMWNO          & 0.0735 $\pm$ 3e-5  & 0.1606 $\pm$2e-4 & 0.0083 $\pm$5e-4 & 0.1139 $\pm$ 3e-4 \\
COMPOL-ATN     & 0.1132 $\pm$ 1e-4  & 0.1523 $\pm$ 3e-4 & 0.0061 $\pm$3e-5  & 0.0888 $\pm$ 2e-4\\
COMPOL-RNN     & 0.1143 $\pm$1e-4 & 0.1627 $\pm$ 7e-5 & 0.0066 $\pm$1e-4 & 0.1136 $\pm$ 1e-4 \\
DiffusionPDE   & 0.0441 $\pm$ 3e-4 & 0.1098 $\pm$ 1e-4 & 0.0026 $\pm$ 1e-4  & 0.0067 $\pm$ 2e-4 \\
CViT-B         & \textbf{0.0163 $\pm$ 6e-4} & \textcolor{blue}{\textbf{0.0759 $\pm$ 6e-5}} & 0.0209 $\pm$2e-3 & 0.1055 $\pm$ 1e-4 \\
\midrule
\textsc{coupled-flow}-Linear & 0.0234 $\pm$ 1e-4 & 0.0798 $\pm$ 3e-5 & \textcolor{blue}{\textbf{0.0023 $\pm$1e-5}} & \textcolor{blue}{\textbf{0.0061 $\pm$ 3e-4}} \\
\textsc{coupled-flow}-GP    & \textcolor{blue}{\textbf{0.0216 $\pm$2e-5}}  & \textbf{0.0731 $\pm$ 7e-5} &  \textbf{0.0019 $\pm$3e-5} & \textbf{0.0052 $\pm$2e-4}  \\
\bottomrule
\end{tabular}
}
\vspace{2mm}
\caption{
Coefficient-shift generalization on Lotka--Volterra and Gray--Scott. 
We report relative $L_2$ errors (mean $\pm$ std) under in-distribution (ID) and out-of-distribution (OOD) coefficient settings.
}
\label{tab:ood_coeff}
\end{table}

\section{Detailed Experiment setup and hyperparameters}
\label{appdix:detailed_exp_setup}
The experiments are conducted with 5 independent runs on the NVIDIA H100.
 The five simulated datasets were stimulated with numerical solvers using 256-point meshes for 1-D cases and $64 \times 64$ meshes for 2-D cases. As prior practice \citep{sun2025compolunifiedneuraloperator}, we also trained the models on two dataset sizes ($512$ and $1024$ samples) and evaluated performance on an independent test set of $200$ samples. Coefficients and boundary conditions remained constant to isolate the effects of varying initial conditions.
The number of parameters for each baseline across different datasets is provided in Tab.~\ref{tab:param_count}. The wall-clock time for each baseline across different datasets is provided in Tab.~\ref{tab:infer_time}. The details of the hyperparameters are provided in Tab~\ref{tab:hyperparameters}.
The official code repository is provided at \url{https://github.com/khiembk/CoupledFlow}.
\section{Limitations}
\label{sec:limitation}
Our method has some limitations that are potential research directions for future work. First, \textsc{coupled-flow} assumes access to paired endpoint data $(\bm{x}, \bm{x}')$, where $\bm{x}'$ denotes the initial coupled state and $\bm{x}'$ denotes the corresponding future solution state. 
This setting is standard in neural operator learning, but it still requires high-fidelity numerical simulations to construct the training dataset. 
In practical scientific applications, however, one may only have sparse trajectories, noisy sensor measurements, or partially observed physical fields. 
Extending \textsc{coupled-flow} to learn from incomplete trajectories or partially observed coupled states is therefore an important direction for future work. 
Potential solutions include combining the proposed coupled transport objective with data-assimilation methods, uncertainty-aware training, or weakly supervised objectives that do not require fully paired endpoint observations. Second, \textsc{coupled-flow} is designed for efficient one-step inference, but its training procedure is more expensive than standard endpoint-regression neural operators. 
A useful future direction is to develop more efficient approximations of the local consistency objective, such as lightweight directional-derivative estimators, reduced-frequency consistency regularization, or shared low-rank coupling structures. 
Such improvements could further improve the scalability of \textsc{coupled-flow} while preserving its one-step inference advantage.

\section{Broader Impacts}
\label{sec:impact}

This work develops a one-step neural operator for coupled multi-physics PDE systems. The main positive impact is that accurate and efficient surrogate models can reduce the cost of repeatedly solving expensive coupled PDEs. 
This can benefit scientific and engineering applications such as subsurface flow modeling, thermal-hydro-mechanical simulation, climate and environmental modeling, materials design, and other domains where high-fidelity numerical simulation is costly. 
Because \textsc{coupled-flow} performs one-step inference after training, it may also reduce inference-time energy consumption compared with iterative generative PDE solvers or repeated numerical simulations.

At the same time, surrogate PDE solvers can have negative consequences if used without proper validation. 
In safety-critical settings, such as geomechanics, energy systems, environmental forecasting, or infrastructure design, inaccurate predictions could lead to incorrect scientific conclusions or poor engineering decisions. 
The proposed method should therefore be viewed as an acceleration tool rather than a replacement for validated numerical solvers. 
For high-stakes applications, users should perform uncertainty quantification, out-of-distribution testing, and domain-specific physical validation before deployment.

The work does not involve human subjects, personal data, private information, or crowdsourced annotations. 
The experiments are conducted on simulated and publicly available scientific benchmark datasets. 
As a result, we do not anticipate direct privacy, fairness, or human-subject risks. 
However, like other scientific machine-learning tools, the method could be applied in domains with environmental or industrial consequences. 
Responsible use requires transparent reporting of model assumptions, data-generation procedures, and failure cases.

\section{Details of Multi-Physics Datasets}
\paragraph{1-D Lotka--Volterra.}
The Lotka--Volterra system models predator-prey dynamics with spatial diffusion. 
It contains two coupled fields $u(x,t)$ and $v(x,t)$:
\begin{eqnarray}
\label{eq:pde_lotka_volterra}
    \frac{\partial u}{\partial t}
    &=&D_u \nabla^2 u+au-buv,
    \nonumber \\
    \frac{\partial v}{\partial t}&=&D_v \nabla^2 v+cuv- dv . \nonumber
\end{eqnarray}
Follow the prior setting~\citep{sun2025compolunifiedneuraloperator}, the initial conditions are sampled from a one-dimensional Gaussian random field with length-scale $l=0.1$ and amplitude $\sigma=1$. The interaction parameters are fixed as $a=b=c=d=0.01$, and the diffusion coefficients are fixed as $D_u=D_v=0.01$. 
The dataset is discretized on a 1-D mesh, and the learning task is to predict the later coupled state $(u,v)$ from the earlier coupled state.

\paragraph{1-D Belousov--Zhabotinsky.}
The Belousov--Zhabotinsky system describes nonlinear chemical oscillations with three coupled reactants $u(x,t)$, $v(x,t)$, and $w(x,t)$:
\begin{eqnarray}
\label{eq:pde_bz}
    \frac{\partial u}{\partial t}&=&
    \epsilon_1 \nabla^2 u+u+v-uv-u^2,\nonumber \\
    \frac{\partial v}{\partial t}&=&\epsilon_2 \nabla^2 v+w-v-uv,\nonumber \\
    \frac{\partial w}{\partial t}&=&\epsilon_3 \nabla^2 w+u-w. \nonumber
\end{eqnarray}
Follow the prior setting~\citep{sun2025compolunifiedneuraloperator}, we set $\epsilon_1=10^{-2}$, $\epsilon_2=10^{-2}$, and $\epsilon_3=5\times 10^{-3}$. 
The system is simulated from $t=0$ to $t=0.5$.

\paragraph{2-D Gray--Scott.}
The Gray--Scott system models two-species reaction-diffusion pattern formation. 
It contains an activator $u(x,y,t)$ and an inhibitor $v(x,y,t)$:
\begin{eqnarray}
\label{eq:pde_grayscott}
    \frac{\partial u}{\partial t}&=&D_u \nabla^2 u-uv^2+F(1-u),\nonumber \\
    \frac{\partial v}{\partial t}&=&D_v \nabla^2 v+uv^2-(F+k)v . \nonumber
\end{eqnarray}
Follow prior practice~\citep{sun2025compolunifiedneuraloperator}, we use the parameter setting $D_u=0.12$, $D_v=0.06$, $F=0.054$, and $k=0.063$. 
The initial state is sampled from a two-dimensional Gaussian random field and evolved until $T=20$. 
The learning task is to predict the future coupled field $(u,v)$ from the earlier field.

\paragraph{2-D multiphase flow.}
The multiphase benchmark describes oil-water two-phase flow in porous media. 
Let $\alpha$ and $\beta$ denote the water and oil components. 
The governing equations follow component-wise mass conservation:
\begin{eqnarray}
\label{eq:pde_multiphase_mass_short}
    \frac{\partial M^\alpha}{\partial t}&=&-
    \nabla\cdot F^\alpha_a+q^\alpha,\nonumber \\
    \frac{\partial M^\beta}{\partial t}
    &=&-\nabla\cdot F^\beta_a+q^\beta, \nonumber
\end{eqnarray}
where $M^\alpha,M^\beta$ are component mass accumulations, $F^\alpha_a,F^\beta_a$ are advective fluxes, and $q^\alpha,q^\beta$ are source or sink terms. 
Following the simulation setup, diffusive fluxes are neglected. 
The phase velocity is given by Darcy's law:
\begin{eqnarray}
\label{eq:pde_multiphase_darcy_short}
    u_p=-k\left(\nabla P_p-\rho_p g\right)\frac{k_{rp}}{\mu_p}, \nonumber
\end{eqnarray}
where $k$ is permeability, $P_p$ is phase pressure, $\rho_p$ is phase density, $k_{rp}$ is relative permeability, and $\mu_p$ is viscosity. Follow \citep{sun2025compolunifiedneuraloperator}, the simulations are performed with GEOS on a $64\times64$ two-dimensional grid. 
Water is injected from two randomly placed sources on the left boundary, and oil is produced from two randomly placed sinks on the right boundary. 
The permeability field satisfies $k_x=k_y$ and is generated by a fractal algorithm with $k_{\min}=1$mD, $k_{\max}=1000$mD, and $k_{\rm base}=100$mD. 
The dataset contains $1024$ scenarios, each simulated for $15$ time steps with step size $5\times10^5$ seconds. 
The stored variables are phase pressure $P_p$ and phase saturation $S_p$, resulting in data of size $1024\times15\times64\times64\times2$.

\paragraph{2-D thermo-hydro-mechanical system.}
The thermo-hydro-mechanical benchmark models coupled pressure diffusion, heat transfer, and mechanical deformation in porous media. 
The governing equations consist of momentum balance, fluid mass balance, and heat transfer:
\begin{eqnarray}
\label{eq:pde_thm}
    G u_{i,jj}&+&\frac{G}{1-2\nu}u_{j,ji}-\alpha p_{,i}-\frac{2G\alpha_m^T(1+\nu)}{3(1-2\nu)}T_{,i}+F_i=0,\nonumber \\
    \frac{\partial p}{\partial t}&=&M\left[\frac{k}{\mu}p_{,jj}-\alpha\frac{\partial \epsilon_{kk}}{\partial t}+\left(\alpha \alpha_m^T+\phi_0(\alpha_f^T-\alpha_m^T)\right)\frac{\partial T}{\partial t}\right]+\gamma,\nonumber \\
    \frac{\partial T}{\partial t}&=&\frac{1}{\rho_t C_t}(\kappa^T T_{,i})_{,i}-\frac{\rho_f C_f}{\rho_t C_t}v_i T_{,i}. \nonumber
\end{eqnarray}
Here, $G$ is the shear modulus, $\nu$ is Poisson's ratio, $\alpha$ is Biot's coefficient, $p$ is pore pressure, $T$ is temperature, $F_i$ is the body force, $M$ is the Biot modulus, $k/\mu$ is hydraulic conductivity, $\epsilon_{kk}$ is volumetric strain, $\phi_0$ is initial porosity, $\alpha_m^T$ and $\alpha_f^T$ are thermal expansion coefficients of porous media and fluid, $\gamma$ is the fluid source or sink term, $\rho_t,C_t$ are total density and heat capacity, $\kappa^T$ is the thermal conductivity tensor, and $\rho_f,C_f$ are fluid density and heat capacity.

\paragraph{2-D Reaction-Diffusion from PDEBench.}
We also evaluate on the public Reaction-Diffusion-2D benchmark from PDEBench. 
This system contains two coupled variables, an activator $u(t,x,y)$ and an inhibitor $v(t,x,y)$:
\begin{eqnarray}
\label{eq:pde_pdebench_reaction_diffusion}
    \frac{\partial u}{\partial t}&=&D_u\left(\frac{\partial^2 u}{\partial x^2}+\frac{\partial^2 u}{\partial y^2}\right)+R_u(u,v),\nonumber \\
    \frac{\partial v}{\partial t}&=&D_v\left(\frac{\partial^2 v}{\partial x^2}+\frac{\partial^2 v}{\partial y^2}\right)+R_v(u,v). \nonumber
\end{eqnarray}
The reaction terms follow the FitzHugh--Nagumo form:
\begin{eqnarray}
\label{eq:pde_pdebench_reaction_terms}
    R_u(u,v)=u-u^3-k-v,\qquad R_v(u,v)=u-v . \nonumber
\end{eqnarray}

The PDEBench release provides trajectories with two variables over time, and we use the same preprocessing and train/test split protocol as described in our experimental setup.

\paragraph{2-D Navier--Stokes from PDEArena.}
We further evaluate on the public Navier-Stokes-2D benchmark from PDEArena. 
This dataset contains a scalar advected field $c(t,x,y)$ and an incompressible velocity field $v(t,x,y)=(v_x(t,x,y),v_y(t,x,y))$, giving three coupled channels $(c,v_x,v_y)$. 
The dynamics can be written as an incompressible buoyancy-driven flow:
\begin{eqnarray}
\label{eq:pde_pdearena_ns}
    \frac{\partial c}{\partial t}+v\cdot\nabla c&=&0,\nonumber \\
    \frac{\partial v}{\partial t}+(v\cdot\nabla)v&=&-\nabla p+\nu \Delta v+c b,\nonumber \\
    \nabla\cdot v&=&0 \nonumber,
\end{eqnarray}
where $p$ is the pressure, $\nu$ is the viscosity, and $b=(0,b_y)$ is the buoyancy direction. 
The PDEArena generator initializes the scalar field with a smooth random noise field and initializes the velocity field to zero. 
At each solver step, the scalar field is advected by the velocity, the velocity is advected by itself, buoyancy forcing is added, viscosity is applied through diffusion, and the velocity is projected to an incompressible field. 
The official HDF5 trajectories store the scalar channel $c$ and the velocity channels $v_x,v_y$, which are used as the three coupled processes in our experiments.

\section{Related Work}\label{sec:related}

\subsection{Flow-based Generative Models}
Generative models based on flow matching \citep{lipman2022flow, lipman2024flow} offer a scalable alternative to diffusion methods by defining a continuous normalizing flows through the means of ordinary differential equations (ODEs). Flow matching transport the samples from a source distribution to a target distribution by learning a time-dependent velocity field using a regression target without the need to simulate during training. During inference, a new sample from the target distribution is generated by integrating the ODEs. Recent advancements aim to reduce the ODE integration costs by straightening flow trajectories such as ReFlow \citep{liu2022flow}, Optimal FLow Matching \citep{kornilov2024optimal}. Other methods such as MeanFlow \citep{geng2025mean} or Consistency Models \cite{song2023consistency} directly learn one-step mappings between the source to target distributions, thus entirely avoiding the costly ODE integration.

\subsection{Single Operator Learning}
\label{sec:related_single_operator}
Neural operator methods, which leverage neural networks as their foundational architecture, present an innovative approach to surrogate modeling that maps input functions to output functions. FNO~\citep{li2021fourier} formulates a neural operator by parameterizing
the integral kernel directly in Fourier space, allowing for an expressive and efficient architecture. Building on these innovations, researchers introduced the GNO \citep{li2020neuraloperatorgraphkernel} and MGNO \citep{li2020multipolegraphneuraloperator}, which combine Nystrom approximation with graph neural networks to approximate function convolution. Another significant contribution emerged with the DeepONet~\citep{lu2021learning}, which employs a dual-network architecture combining a branch network for input
functions and a trunk network for sampling locations. 
Recently, \cite{wang2025cvitcontinuousvisiontransformer} propose CVIT, which combines a vision transformer encoder, a novel grid-based coordinate embedding, and a query-wise cross-attention mechanism to capture multi-scale dependencies effectively.

Generative models have gained increasing interests in recent years to solve PDEs. DiffusionPDE \citep{huang2024diffusionpde} uses denoising diffusion probabilistic models to jointly model the distributions of the coefficient and solution spaces in order to solve forward as well as backward PDE problems under missing data regimes. DYffusion \citep{ruhling2023dyffusion} leverages dynamics of the spatiotemporal data to train diffusion models, facilitating multi-step and long-range PDE predictions. Physics-Based Flow Matching (PBFM) \citep{baldan2025flow} and Physics-Informed Diffusion Models (PBDM) \citep{bastek2024physics} combine generative loss and physical residual in the objective functions to learn the underlying physics of the data. ECI \citep{cheng2024gradient} and PCFM \citep{utkarsh2025physics} incorporate additional projection step to the generated solution samples to enforce hard physical constraints. Although these methods have shown great potential in modeling single-physics systems, extending them to multiphysics systems remains a non-trivial challenge , requiring further advances in modeling cross-physics interactions and capturing complex coupling dynamics.

\subsection{Coupled Operator Learning}
\label{sec:related_coupled_operator}


Coupled operator learning extends standard neural operator learning from single-field systems to multi-physics systems in which several physical variables evolve jointly and influence one another. Rather than learning a single operator from one input field to one output field, the goal is to learn a family of operators whose predictions for each process depend on the full coupled state. CMWNO~\citep{xiao2023coupled} addresses this problem by decomposing coupled integral kernels in the multiwavelet domain, allowing cross-process interactions to be represented during wavelet decomposition and reconstruction. While effective, this coupling mechanism is tied to a particular wavelet-space factorization and predefined decomposition structure, which may limit flexibility when the number of processes, coupling patterns, or spatial resolutions varies. More recently,  COMPOL~\citep{sun2025compolunifiedneuraloperator} builds on Fourier neural operators by maintaining process-specific operator streams and exchanging information through recurrent or attention-based aggregation modules. This architecture provides a flexible way to model inter-process dependencies in latent feature space and has shown strong empirical performance on several coupled PDE datasets. Related attention-based neural operators also suggest that explicit information exchange across physical variables can improve multi-physics prediction.  However, these approaches remain primarily \emph{architecture-centric}: they model coupling through feature aggregation inside the neural operator, while the learning objective is still based on direct endpoint regression from initial states to future states.

\newpage
\section*{NeurIPS Paper Checklist}

\begin{enumerate}

\item {\bf Claims}
    \item[] Question: Do the main claims made in the abstract and introduction accurately reflect the paper's contributions and scope?
    \item[] Answer: \answerYes{}
    \item[] Justification: see Section~\ref{sec:method}

\item {\bf Limitations}
    \item[] Question: Does the paper discuss the limitations of the work performed by the authors?
    \item[] Answer: \answerYes{} 
    \item[] Justification: see Appendix~\ref{sec:limitation}

\item {\bf Theory assumptions and proofs}
    \item[] Question: For each theoretical result, does the paper provide the full set of assumptions and a complete (and correct) proof?
    \item[] Answer: \answerYes{} 
    \item[] Justification: see Appendix~\ref{appendix:proof_thm_1} and Appendix~\ref{appendix:proof_theorem_2}

    \item {\bf Experimental result reproducibility}
    \item[] Question: Does the paper fully disclose all the information needed to reproduce the main experimental results of the paper to the extent that it affects the main claims and/or conclusions of the paper (regardless of whether the code and data are provided or not)?
    \item[] Answer: \answerYes{}{} 
    \item[] Justification: see Appendix~\ref{appdix:detailed_exp_setup}

\item {\bf Open access to data and code}
    \item[] Question: Does the paper provide open access to the data and code, with sufficient instructions to faithfully reproduce the main experimental results, as described in supplemental material?
    \item[] Answer: \answerYes{} 
    \item[] Justification: see Appendix~\ref{appdix:detailed_exp_setup} 

\item {\bf Experimental setting/details}
    \item[] Question: Does the paper specify all the training and test details (e.g., data splits, hyperparameters, how they were chosen, type of optimizer) necessary to understand the results?
    \item[] Answer: \answerYes{} 
    \item[] Justification: see Section~\ref{sec:experiments} and Appendix~\ref{appdix:detailed_exp_setup}

\item {\bf Experiment statistical significance}
    \item[] Question: Does the paper report error bars suitably and correctly defined or other appropriate information about the statistical significance of the experiments?
    \item[] Answer: \answerYes{} 
    \item[] Justification: see Appendix~\ref{appendix:Experiment_detailed_results}

\item {\bf Experiments compute resources}
    \item[] Question: For each experiment, does the paper provide sufficient information on the computer resources (type of compute workers, memory, time of execution) needed to reproduce the experiments?
    \item[] Answer: \answerYes{} 
    \item[] Justification: see Appendix~\ref{appdix:detailed_exp_setup}
    
\item {\bf Code of ethics}
    \item[] Question: Does the research conducted in the paper conform, in every respect, with the NeurIPS Code of Ethics \url{https://neurips.cc/public/EthicsGuidelines}?
    \item[] Answer: \answerYes{} 
    \item[] Justification: The work focuses on learning surrogate neural operators for coupled multi-physics PDE systems using simulated and publicly available scientific benchmarks. It does not involve human subjects, personally identifiable information, sensitive data, or deployment in safety-critical real-world settings. We have cited the datasets and prior methods used in the experiments and followed standard scientific reporting practices.

\item {\bf Broader impacts}
    \item[] Question: Does the paper discuss both potential positive societal impacts and negative societal impacts of the work performed?
    \item[] Answer: \answerYes{} 
    \item[] Justification: see Appendix~\ref{sec:impact}
    
\item {\bf Safeguards}
    \item[] Question: Does the paper describe safeguards that have been put in place for responsible release of data or models that have a high risk for misuse (e.g., pre-trained language models, image generators, or scraped datasets)?
    \item[] Answer: \answerNA{} 
    \item[] Justification: The paper does not release data or models with a high risk for misuse. The work focuses on surrogate modeling for coupled multi-physics PDE systems using simulated and publicly available scientific benchmark datasets, and does not involve scraped datasets, human-subject data, large pretrained generative models, or systems intended for safety-critical deployment.

\item {\bf Licenses for existing assets}
    \item[] Question: Are the creators or original owners of assets (e.g., code, data, models), used in the paper, properly credited and are the license and terms of use explicitly mentioned and properly respected?
    \item[] Answer: \answerYes{}.
    \item[] Justification: We use existing public benchmark datasets and baseline implementations only for research and reproducibility purposes. The original creators are credited through citations to the corresponding papers and official repositories. For each dataset and codebase used in our experiments, we report the source, version or access date when applicable. We respect the stated terms of use and do not repackage or redistribute existing assets beyond what is allowed by their licenses.

\item {\bf New assets}
    \item[] Question: Are new assets introduced in the paper well documented and is the documentation provided alongside the assets?
    \item[] Answer: \answerYes{} 
    \item[] Justification: We will release the implementation of our method, including training and inference scripts, configuration files, and instructions for reproducing the reported experiments. The released assets will be documented with usage instructions, required dependencies, hyperparameter settings, and limitations. Our work does not introduce a new dataset involving human subjects, and all benchmark datasets used in the paper are existing public datasets. The release will be anonymized during the review period.

\item {\bf Crowdsourcing and research with human subjects}
    \item[] Question: For crowdsourcing experiments and research with human subjects, does the paper include the full text of instructions given to participants and screenshots, if applicable, as well as details about compensation (if any)? 
    \item[] Answer: \answerNA{} 
    \item[] Justification: The paper does not involve crowdsourcing, user studies, or research with human subjects. All experiments are conducted on simulated and publicly available PDE benchmark datasets, so there are no participant instructions, screenshots, or compensation details to report.

\item {\bf Institutional review board (IRB) approvals or equivalent for research with human subjects}
    \item[] Question: Does the paper describe potential risks incurred by study participants, whether such risks were disclosed to the subjects, and whether Institutional Review Board (IRB) approvals (or an equivalent approval/review based on the requirements of your country or institution) were obtained?
    \item[] Answer: \answerNA{} 
    \item[] Justification: The paper does not involve research with human subjects, crowdsourcing, user studies, personally identifiable information, or participant data. All experiments are conducted on simulated and publicly available PDE benchmark datasets, so IRB approval or equivalent human-subjects review is not applicable.

\item {\bf Declaration of LLM usage}
    \item[] Question: Does the paper describe the usage of LLMs if it is an important, original, or non-standard component of the core methods in this research? Note that if the LLM is used only for writing, editing, or formatting purposes and does \emph{not} impact the core methodology, scientific rigor, or originality of the research, declaration is not required.
    \item[] Answer: \answerNA{} 
    \item[] Justification: LLM is used only for writing, editing, or formatting purposes and does not impact the core methodology, scientific rigor, or originality of the research.

\end{enumerate}

\end{document}